\pdfoutput=1

\documentclass[11pt]{article}
\usepackage{xcolor} %

\newcommand{\tbd}{\textcolor{gray}{tbd}}
\newcommand{\rv}[1]{\ifcsname rv@#1\endcsname\csname rv@#1\endcsname\else\tbd\fi}
\expandafter\def\csname rv@abj.head.infl\endcsname{35.53}
\expandafter\def\csname rv@abj.head.l3\endcsname{72.00}
\expandafter\def\csname rv@abj.head.subst\endcsname{31.19}
\expandafter\def\csname rv@abj.ours.delta\endcsname{+3.39 (5/5)}
\expandafter\def\csname rv@abj.ours.infl\endcsname{76.52}
\expandafter\def\csname rv@abj.ours.l3\endcsname{84.44}
\expandafter\def\csname rv@abj.ours.subst\endcsname{51.26}
\expandafter\def\csname rv@abj.rule.delta\endcsname{+9.05 (5/5)}
\expandafter\def\csname rv@abj.rule.infl\endcsname{68.59}
\expandafter\def\csname rv@abj.rule.l3\endcsname{81.04}
\expandafter\def\csname rv@abj.rule.subst\endcsname{42.81}
\expandafter\def\csname rv@abl.evidence.delta\endcsname{+1.06 (4/5)}
\expandafter\def\csname rv@abl.evidence.infl\endcsname{77.87}
\expandafter\def\csname rv@abl.evidence.l3\endcsname{84.24}
\expandafter\def\csname rv@abl.evidence.subst\endcsname{49.06}
\expandafter\def\csname rv@abl.gate.delta\endcsname{+0.94 (5/5)}
\expandafter\def\csname rv@abl.gate.infl\endcsname{78.36}
\expandafter\def\csname rv@abl.gate.l3\endcsname{85.37}
\expandafter\def\csname rv@abl.gate.subst\endcsname{53.60}
\expandafter\def\csname rv@abl.plain.infl\endcsname{74.59}
\expandafter\def\csname rv@abl.plain.l3\endcsname{83.18}
\expandafter\def\csname rv@abl.plain.subst\endcsname{47.47}
\expandafter\def\csname rv@abl.split.delta\endcsname{+0.23 (3/5)}
\expandafter\def\csname rv@abl.stretch.delta\endcsname{+0.19 (3/5)}
\expandafter\def\csname rv@abl.stretch.infl\endcsname{76.52}
\expandafter\def\csname rv@abl.stretch.l3\endcsname{84.44}
\expandafter\def\csname rv@abl.stretch.subst\endcsname{51.26}
\expandafter\def\csname rv@ac.copy_floor.f\endcsname{69.04}
\expandafter\def\csname rv@ac.copy_floor.frame\endcsname{0.00}
\expandafter\def\csname rv@ac.copy_floor.infl\endcsname{0.00}
\expandafter\def\csname rv@ac.copy_floor.inv\endcsname{0.66}
\expandafter\def\csname rv@ac.copy_floor.l3\endcsname{53.47}
\expandafter\def\csname rv@ac.copy_floor.p\endcsname{80.73}
\expandafter\def\csname rv@ac.copy_floor.r\endcsname{60.32}
\expandafter\def\csname rv@ac.copy_floor.subst\endcsname{0.00}
\expandafter\def\csname rv@ac.em_aligner.f\endcsname{71.87}
\expandafter\def\csname rv@ac.em_aligner.frame\endcsname{8.46}
\expandafter\def\csname rv@ac.em_aligner.infl\endcsname{66.07}
\expandafter\def\csname rv@ac.em_aligner.inv\endcsname{1.15}
\expandafter\def\csname rv@ac.em_aligner.l3\endcsname{67.46}
\expandafter\def\csname rv@ac.em_aligner.p\endcsname{84.87}
\expandafter\def\csname rv@ac.em_aligner.r\endcsname{62.39}
\expandafter\def\csname rv@ac.em_aligner.subst\endcsname{1.32}
\expandafter\def\csname rv@ac.gold_links.f\endcsname{100.00}
\expandafter\def\csname rv@ac.gold_links.frame\endcsname{100.00}
\expandafter\def\csname rv@ac.gold_links.infl\endcsname{82.61}
\expandafter\def\csname rv@ac.gold_links.inv\endcsname{0.00}
\expandafter\def\csname rv@ac.gold_links.l3\endcsname{91.74}
\expandafter\def\csname rv@ac.gold_links.p\endcsname{100.00}
\expandafter\def\csname rv@ac.gold_links.r\endcsname{100.00}
\expandafter\def\csname rv@ac.gold_links.subst\endcsname{78.03}
\expandafter\def\csname rv@ac.nmt_aligner.f\endcsname{66.66}
\expandafter\def\csname rv@ac.nmt_aligner.frame\endcsname{10.02}
\expandafter\def\csname rv@ac.nmt_aligner.infl\endcsname{33.80}
\expandafter\def\csname rv@ac.nmt_aligner.inv\endcsname{0.48}
\expandafter\def\csname rv@ac.nmt_aligner.l3\endcsname{61.25}
\expandafter\def\csname rv@ac.nmt_aligner.p\endcsname{83.50}
\expandafter\def\csname rv@ac.nmt_aligner.r\endcsname{55.56}
\expandafter\def\csname rv@ac.nmt_aligner.subst\endcsname{5.80}
\expandafter\def\csname rv@ac.ours.f\endcsname{94.27}
\expandafter\def\csname rv@ac.ours.frame\endcsname{\textbf{46.68}}
\expandafter\def\csname rv@ac.ours.infl\endcsname{\textbf{78.36}}
\expandafter\def\csname rv@ac.ours.inv\endcsname{0.18}
\expandafter\def\csname rv@ac.ours.l3\endcsname{\textbf{85.37}}
\expandafter\def\csname rv@ac.ours.p\endcsname{94.03}
\expandafter\def\csname rv@ac.ours.r\endcsname{94.54}
\expandafter\def\csname rv@ac.ours.subst\endcsname{\textbf{53.60}}
\expandafter\def\csname rv@ac.ours_plus_span_aligner.f\endcsname{\textbf{95.47}}
\expandafter\def\csname rv@ac.ours_plus_span_aligner.frame\endcsname{16.67}
\expandafter\def\csname rv@ac.ours_plus_span_aligner.infl\endcsname{71.64}
\expandafter\def\csname rv@ac.ours_plus_span_aligner.inv\endcsname{0.14}
\expandafter\def\csname rv@ac.ours_plus_span_aligner.l3\endcsname{83.53}
\expandafter\def\csname rv@ac.ours_plus_span_aligner.p\endcsname{\textbf{96.05}}
\expandafter\def\csname rv@ac.ours_plus_span_aligner.r\endcsname{\textbf{94.91}}
\expandafter\def\csname rv@ac.ours_plus_span_aligner.subst\endcsname{50.41}
\expandafter\def\csname rv@ac.plain_pointer.f\endcsname{94.07}
\expandafter\def\csname rv@ac.plain_pointer.frame\endcsname{37.27}
\expandafter\def\csname rv@ac.plain_pointer.infl\endcsname{74.59}
\expandafter\def\csname rv@ac.plain_pointer.inv\endcsname{0.19}
\expandafter\def\csname rv@ac.plain_pointer.l3\endcsname{83.18}
\expandafter\def\csname rv@ac.plain_pointer.p\endcsname{94.14}
\expandafter\def\csname rv@ac.plain_pointer.r\endcsname{94.06}
\expandafter\def\csname rv@ac.plain_pointer.subst\endcsname{47.47}
\expandafter\def\csname rv@ac.sim_aligner.f\endcsname{82.16}
\expandafter\def\csname rv@ac.sim_aligner.frame\endcsname{16.34}
\expandafter\def\csname rv@ac.sim_aligner.infl\endcsname{25.32}
\expandafter\def\csname rv@ac.sim_aligner.inv\endcsname{0.17}
\expandafter\def\csname rv@ac.sim_aligner.l3\endcsname{66.47}
\expandafter\def\csname rv@ac.sim_aligner.p\endcsname{90.73}
\expandafter\def\csname rv@ac.sim_aligner.r\endcsname{75.17}
\expandafter\def\csname rv@ac.sim_aligner.subst\endcsname{23.21}
\expandafter\def\csname rv@ac.sim_aligner_ft.f\endcsname{86.75}
\expandafter\def\csname rv@ac.sim_aligner_ft.frame\endcsname{14.67}
\expandafter\def\csname rv@ac.sim_aligner_ft.infl\endcsname{37.23}
\expandafter\def\csname rv@ac.sim_aligner_ft.inv\endcsname{0.41}
\expandafter\def\csname rv@ac.sim_aligner_ft.l3\endcsname{67.50}
\expandafter\def\csname rv@ac.sim_aligner_ft.p\endcsname{89.57}
\expandafter\def\csname rv@ac.sim_aligner_ft.r\endcsname{84.37}
\expandafter\def\csname rv@ac.sim_aligner_ft.subst\endcsname{13.05}
\expandafter\def\csname rv@ac.span_aligner.f\endcsname{94.48}
\expandafter\def\csname rv@ac.span_aligner.frame\endcsname{15.18}
\expandafter\def\csname rv@ac.span_aligner.infl\endcsname{67.67}
\expandafter\def\csname rv@ac.span_aligner.inv\endcsname{\textbf{0.14}}
\expandafter\def\csname rv@ac.span_aligner.l3\endcsname{82.02}
\expandafter\def\csname rv@ac.span_aligner.p\endcsname{95.86}
\expandafter\def\csname rv@ac.span_aligner.r\endcsname{93.16}
\expandafter\def\csname rv@ac.span_aligner.subst\endcsname{47.49}
\expandafter\def\csname rv@ac.span_pair.f\endcsname{86.14}
\expandafter\def\csname rv@ac.span_pair.frame\endcsname{43.88}
\expandafter\def\csname rv@ac.span_pair.infl\endcsname{49.02}
\expandafter\def\csname rv@ac.span_pair.inv\endcsname{0.16}
\expandafter\def\csname rv@ac.span_pair.l3\endcsname{68.62}
\expandafter\def\csname rv@ac.span_pair.p\endcsname{91.21}
\expandafter\def\csname rv@ac.span_pair.r\endcsname{81.61}
\expandafter\def\csname rv@ac.span_pair.subst\endcsname{7.70}
\expandafter\def\csname rv@ac.sultan.f\endcsname{88.15}
\expandafter\def\csname rv@ac.sultan.frame\endcsname{12.47}
\expandafter\def\csname rv@ac.sultan.infl\endcsname{56.88}
\expandafter\def\csname rv@ac.sultan.inv\endcsname{0.20}
\expandafter\def\csname rv@ac.sultan.l3\endcsname{70.87}
\expandafter\def\csname rv@ac.sultan.p\endcsname{94.16}
\expandafter\def\csname rv@ac.sultan.r\endcsname{83.07}
\expandafter\def\csname rv@ac.sultan.subst\endcsname{7.17}
\expandafter\def\csname rv@al.A.random.200\endcsname{83.54}
\expandafter\def\csname rv@al.A.uncertainty.200\endcsname{84.58}
\expandafter\def\csname rv@al.all.syn\endcsname{84.44}
\expandafter\def\csname rv@bnd.both.infl\endcsname{91.73}
\expandafter\def\csname rv@bnd.both.l3\endcsname{93.57}
\expandafter\def\csname rv@bnd.both.subst\endcsname{79.63}
\expandafter\def\csname rv@bnd.gold.infl\endcsname{86.89}
\expandafter\def\csname rv@bnd.gold.l3\endcsname{93.09}
\expandafter\def\csname rv@bnd.gold.subst\endcsname{80.33}
\expandafter\def\csname rv@bnd.null.infl\endcsname{87.02}
\expandafter\def\csname rv@bnd.null.l3\endcsname{90.95}
\expandafter\def\csname rv@bnd.null.subst\endcsname{71.56}
\expandafter\def\csname rv@bnd.pred.infl\endcsname{82.61}
\expandafter\def\csname rv@bnd.pred.l3\endcsname{87.32}
\expandafter\def\csname rv@bnd.pred.subst\endcsname{59.09}
\expandafter\def\csname rv@bnd.typing.infl\endcsname{87.14}
\expandafter\def\csname rv@bnd.typing.l3\endcsname{89.61}
\expandafter\def\csname rv@bnd.typing.subst\endcsname{65.65}
\expandafter\def\csname rv@ds.em_aligner.f1\endcsname{49.31}
\expandafter\def\csname rv@ds.em_aligner.r10\endcsname{86.88}
\expandafter\def\csname rv@ds.em_aligner.type\endcsname{78.43}
\expandafter\def\csname rv@ds.gold.type\endcsname{79.62}
\expandafter\def\csname rv@ds.lengths.f1\endcsname{3.51}
\expandafter\def\csname rv@ds.lengths.r10\endcsname{16.07}
\expandafter\def\csname rv@ds.lengths.type\endcsname{36.58}
\expandafter\def\csname rv@ds.ours.f1\endcsname{69.85}
\expandafter\def\csname rv@ds.ours.r10\endcsname{93.99}
\expandafter\def\csname rv@ds.ours.type\endcsname{77.77}
\expandafter\def\csname rv@ds.sim_aligner.f1\endcsname{52.67}
\expandafter\def\csname rv@ds.sim_aligner.r10\endcsname{80.41}
\expandafter\def\csname rv@ds.sim_aligner.type\endcsname{71.09}
\expandafter\def\csname rv@ds.sim_aligner_ft.f1\endcsname{56.28}
\expandafter\def\csname rv@ds.sim_aligner_ft.r10\endcsname{87.39}
\expandafter\def\csname rv@ds.sim_aligner_ft.type\endcsname{74.27}
\expandafter\def\csname rv@ds.similarity.f1\endcsname{51.34}
\expandafter\def\csname rv@ds.similarity.r10\endcsname{89.55}
\expandafter\def\csname rv@ds.span_aligner.f1\endcsname{68.98}
\expandafter\def\csname rv@ds.span_aligner.r10\endcsname{94.02}
\expandafter\def\csname rv@ds.span_aligner.type\endcsname{78.27}
\expandafter\def\csname rv@err.fa.copy.infl\endcsname{13}
\expandafter\def\csname rv@err.fa.copy.subst\endcsname{21}
\expandafter\def\csname rv@err.fa.invented.infl\endcsname{14}
\expandafter\def\csname rv@err.fa.invented.subst\endcsname{23}
\expandafter\def\csname rv@err.fa.other.infl\endcsname{4}
\expandafter\def\csname rv@err.fa.other.subst\endcsname{6}
\expandafter\def\csname rv@err.fa.source.infl\endcsname{1}
\expandafter\def\csname rv@err.fa.source.subst\endcsname{3}
\expandafter\def\csname rv@err.miss.label.infl\endcsname{7}
\expandafter\def\csname rv@err.miss.label.subst\endcsname{2}
\expandafter\def\csname rv@err.miss.nolink.infl\endcsname{19}
\expandafter\def\csname rv@err.miss.nolink.subst\endcsname{18}
\expandafter\def\csname rv@err.miss.source.infl\endcsname{2}
\expandafter\def\csname rv@err.miss.source.subst\endcsname{1}

\newcommand{\AnalysisCoefBlocks}{%
\begin{tabular*}{\textwidth}{@{\extracolsep{\fill}}l*{12}{c}@{}}%
\toprule%
 & \multicolumn{6}{c}{\textbf{Amount}} & \multicolumn{6}{c}{\textbf{Copying}} \\%
\cmidrule(lr){2-7}\cmidrule(lr){8-13}%
 & \stk{reuse\\words} & \stk{source\\words} & \stk{links} & \stk{link\\rate} & \stk{source\\coverage} & \stk{length\\ratio} & \stk{longest\\run} & \stk{runs} & \stk{\texttt{COPY}\\share} & \stk{longest\\quotation} & \stk{quota-\\tions} & \stk{quoted\\share} \\%
\midrule%
\stkl{Reference\\vs.\ no match} & $+0.12$ & $-0.08$ & \cellcolor{shadepos!56}$+2.22$ & $0.00$ & $+0.06$$^{\dagger}$ & $+0.05$$^{\dagger}$ & \cellcolor{shadepos!39}$+1.54$ & \cellcolor{shadepos!9}$+0.36$ & \cellcolor{shadeneg!55}$-2.16$ & \cellcolor{shadepos!6}$+0.23$ & \cellcolor{shadepos!6}$+0.26$ & \cellcolor{shadepos!6}$+0.23$ \\%
\stkl{Verbatim\\vs.\ no match} & \cellcolor{shadepos!7}$+0.33$ & $0.00$ & \cellcolor{shadepos!70}$+3.50$ & $+0.13$ & \cellcolor{shadepos!7}$+0.35$ & $-0.16$ & \cellcolor{shadepos!55}$+2.75$ & \cellcolor{shadepos!10}$+0.52$ & \cellcolor{shadeneg!45}$-2.23$ & $+0.12$ & $+0.13$ & $+0.11$ \\%
\stkl{Allusion\\vs.\ no match} & $+0.06$$^{\dagger}$ & $-0.04$$^{\dagger}$ & \cellcolor{shadepos!37}$+1.21$ & $-0.03$ & $0.00$ & $+0.09$$^{\dagger}$ & \cellcolor{shadepos!6}$+0.19$ & $-0.05$$^{\dagger}$ & \cellcolor{shadeneg!40}$-1.31$ & $+0.04$ & $+0.04$ & $+0.04$ \\%
\stkl{Verbatim\\vs.\ allusion} & \cellcolor{shadepos!7}$+0.22$ & \cellcolor{shadepos!4}$+0.12$ & \cellcolor{shadepos!43}$+1.34$ & \cellcolor{shadepos!11}$+0.33$ & \cellcolor{shadepos!8}$+0.27$ & $-0.05$$^{\dagger}$ & \cellcolor{shadepos!70}$+2.21$ & \cellcolor{shadepos!8}$+0.26$ & \cellcolor{shadeneg!27}$-0.85$ & \cellcolor{shadepos!4}$+0.13$ & \cellcolor{shadeneg!7}$-0.21$$^{\dagger}$ & \cellcolor{shadeneg!7}$-0.24$ \\%
\bottomrule%
\end{tabular*}%
\par\vspace{4pt}%
\begin{tabular*}{\textwidth}{@{\extracolsep{\fill}}l*{11}{c}@{}}%
\toprule%
 & \multicolumn{6}{c}{\textbf{Change}} & \multicolumn{5}{c}{\textbf{Pattern}} \\%
\cmidrule(lr){2-7}\cmidrule(lr){8-12}%
 & \stk{\texttt{INFLECT}\\share} & \stk{\texttt{SUBST}\\unrelated} & \stk{\texttt{SUBST}\\related} & \stk{\texttt{SPLIT}\\\texttt{MERGE}} & \stk{unlinked\\share} & \stk{unclaimed\\share} & \stk{crossing\\links} & \stk{crossing\\share} & \stk{adapted\\phrases} & \stk{disper-\\sals} & \stk{\texttt{MARK}\\words} \\%
\midrule%
\stkl{Reference\\vs.\ no match} & \cellcolor{shadeneg!53}$-2.10$ & \cellcolor{shadeneg!20}$-0.79$ & \cellcolor{shadeneg!21}$-0.81$ & \cellcolor{shadeneg!8}$-0.33$ & $0.00$ & $-0.11$ & \cellcolor{shadeneg!10}$-0.39$ & \cellcolor{shadepos!6}$+0.25$ & $+0.07$ & $-0.03$ & \cellcolor{shadeneg!12}$-0.46$ \\%
\stkl{Verbatim\\vs.\ no match} & \cellcolor{shadeneg!41}$-2.03$ & \cellcolor{shadeneg!17}$-0.85$ & \cellcolor{shadeneg!16}$-0.78$ & \cellcolor{shadeneg!4}$-0.22$ & $-0.13$ & \cellcolor{shadeneg!7}$-0.36$ & \cellcolor{shadeneg!13}$-0.65$ & \cellcolor{shadepos!8}$+0.40$ & $+0.06$$^{\dagger}$ & $-0.08$ & \cellcolor{shadeneg!9}$-0.45$ \\%
\stkl{Allusion\\vs.\ no match} & \cellcolor{shadeneg!45}$-1.46$ & \cellcolor{shadeneg!18}$-0.58$ & \cellcolor{shadeneg!18}$-0.59$ & \cellcolor{shadeneg!8}$-0.25$ & $+0.03$ & $-0.07$$^{\dagger}$ & \cellcolor{shadeneg!8}$-0.26$$^{\dagger}$ & \cellcolor{shadepos!6}$+0.20$$^{\dagger}$ & $0.00$ & $+0.02$$^{\dagger}$ & \cellcolor{shadeneg!10}$-0.31$ \\%
\stkl{Verbatim\\vs.\ allusion} & \cellcolor{shadeneg!28}$-0.90$ & \cellcolor{shadeneg!13}$-0.41$ & \cellcolor{shadeneg!13}$-0.42$ & $0.00$ & \cellcolor{shadeneg!11}$-0.33$ & \cellcolor{shadeneg!8}$-0.26$ & \cellcolor{shadeneg!23}$-0.73$ & \cellcolor{shadepos!16}$+0.50$ & $+0.09$ & \cellcolor{shadeneg!6}$-0.19$ & \cellcolor{shadeneg!4}$-0.12$$^{\dagger}$ \\%
\bottomrule%
\end{tabular*}%
\par\vspace{4pt}%
\begin{tabular*}{\textwidth}{@{\extracolsep{\fill}}l*{7}{c}@{}}%
\toprule%
 & \multicolumn{7}{c}{\textbf{Model}} \\%
\cmidrule(lr){2-8}%
 & \stk{mean link\\conf.} & \stk{min link\\conf.} & \stk{no-source\\conf.} & \stk{verbatim\\assoc.} & \stk{allusion\\assoc.} & \stk{marker\\assoc.} & \stk{no-match\\assoc.} \\%
\midrule%
\stkl{Reference\\vs.\ no match} & \cellcolor{shadepos!70}$+2.77$ & \cellcolor{shadepos!16}$+0.62$ & $+0.04$ & $0.00$ & $-0.03$$^{\dagger}$ & $+0.01$$^{\dagger}$ & $0.00$ \\%
\stkl{Verbatim\\vs.\ no match} & \cellcolor{shadepos!33}$+1.65$ & \cellcolor{shadepos!33}$+1.66$ & $+0.05$ & $+0.17$ & \cellcolor{shadeneg!4}$-0.18$ & \cellcolor{shadeneg!4}$-0.21$$^{\dagger}$ & $-0.06$$^{\dagger}$ \\%
\stkl{Allusion\\vs.\ no match} & \cellcolor{shadepos!70}$+2.27$ & \cellcolor{shadepos!13}$+0.41$ & $+0.03$$^{\dagger}$ & $-0.04$ & $+0.02$$^{\dagger}$ & $-0.02$$^{\dagger}$ & $+0.03$$^{\dagger}$ \\%
\stkl{Verbatim\\vs.\ allusion} & $+0.10$$^{\dagger}$ & \cellcolor{shadepos!13}$+0.43$ & $0.00$ & \cellcolor{shadepos!13}$+0.40$ & \cellcolor{shadeneg!11}$-0.34$ & \cellcolor{shadepos!7}$+0.21$$^{\dagger}$ & \cellcolor{shadeneg!11}$-0.35$ \\%
\bottomrule%
\end{tabular*}%
}
\newcommand{\AnalysisFeatureRows}{%
\texttt{n\_q} & words in the reusing passage & amount & 0.47 \\%
\texttt{n\_s} & words in the source passage & amount & 0.48 \\%
\texttt{n\_links} & links drawn & amount & 0.85 \\%
\texttt{link\_rate} & links per reuse word & amount & 0.81 \\%
\texttt{src\_cov} & links per source word & amount & 0.79 \\%
\texttt{len\_ratio} & reuse length over source length & amount & 0.49 \\%
\texttt{longest\_run} & longest run of consecutive links & copying & 0.84 \\%
\texttt{n\_runs} & number of linked runs & copying & 0.54 \\%
\texttt{copy\_share} & share of links that are copies & copying & 0.69 \\%
\texttt{quote\_longest} & longest verbatim quotation, in words & copying & 0.68 \\%
\texttt{quote\_n} & number of verbatim quotations & copying & 0.68 \\%
\texttt{quote\_share} & share of reuse words inside a verbatim quotation & copying & 0.68 \\%
\texttt{inflect\_share} & share of links that are inflections & change & 0.41 \\%
\texttt{unrelated\_subst\_share} & share of links that are unrelated substitutions (no named relation) & change & 0.53 \\%
\texttt{related\_subst\_share} & share of links that are related substitutions (a named relation) & change & 0.48 \\%
\texttt{split\_merge\_share} & share of links that are splits or merges & change & 0.50 \\%
\texttt{unlinked\_share} & share of reuse words without a link & change & 0.19 \\%
\texttt{unclaimed\_share} & share of source words without a link & change & 0.21 \\%
\texttt{reorder\_n} & links that cross another link & pattern & 0.52 \\%
\texttt{reorder\_share} & share of links that cross another link & pattern & 0.51 \\%
\texttt{adapt\_n} & reused phrases with substitutions concentrated inside them & pattern & 0.51 \\%
\texttt{disperse\_n} & shared material split across a clause boundary & pattern & 0.50 \\%
\texttt{mark\_n} & words of a citing formula & pattern & 0.58 \\%
\texttt{mean\_link\_p} & mean link probability of the drawn links & model & 0.73 \\%
\texttt{min\_link\_p} & lowest link probability among the drawn links & model & 0.68 \\%
\texttt{mean\_null\_p\_unlinked} & mean no-source probability of the unlinked reuse words & model & 0.50 \\%
\texttt{assoc\_verbatim} & share of reuse words with verbatim association & model & 0.82 \\%
\texttt{assoc\_allusion} & share of reuse words with allusion association & model & 0.53 \\%
\texttt{assoc\_marker} & share of reuse words with marker association & model & 0.58 \\%
\texttt{assoc\_nomatch} & share of reuse words with no-match association & model & 0.19 \\%
}
\expandafter\def\csname rv@ana.auc.lex-rel-share\endcsname{0.48}
\expandafter\def\csname rv@ana.auc.subst-share\endcsname{0.53}
\expandafter\def\csname rv@ana.features\endcsname{30}
\expandafter\def\csname rv@ana.negperpos\endcsname{10}

\expandafter\def\csname rv@compute.hours.reported\endcsname{220}
\expandafter\def\csname rv@compute.hours.tuning\endcsname{18}
\expandafter\def\csname rv@compute.params.encoder\endcsname{111.3}
\expandafter\def\csname rv@compute.params.heads\endcsname{4.4}

\def\DataAllOpsSlimRows{%
\multicolumn{9}{@{}l}{\emph{Links (share of links)}} \\
\opTok{COPY} & form & verb. & same form, spelling folded & \textit{frigidus} / \textit{frigidus} & 4,828 & 77.2 & 85.2 & 48.0 \\
\opTok{INFLECT} & form & verb. & same lemma, other form & \textit{regem} / \textit{reges} & 870 & 13.9 & 9.3 & 30.8 \\
\opTok{SUBST} & lexical & allus. & other lemma, same slot & \textit{pugnam} / \textit{calamitatem} & 503 & 8.0 & 4.7 & 20.5 \\
\opTok{SPLIT} & cardinality & allus. & two reuse words, one source & n/a & 40 & 0.6 & 0.7 & 0.4 \\
\opTok{MERGE} & cardinality & allus. & one reuse word, two sources & n/a & 9 & 0.1 & 0.1 & 0.2 \\
\multicolumn{9}{@{}l}{\quad \emph{Relations of \opTok{SUBST} (share of \opTok{SUBST} links)}} \\
\quad \opTok{SYN} & lexical & allus. & same sense, other lemma & \textit{tum} / \textit{tunc} & 199 & 39.6 & 33.2 & 44.9 \\
\quad \opTok{DERIV} & lexical & allus. & derivational relative & \textit{luxuriem} / \textit{luxuriant} & 47 & 9.3 & 8.3 & 10.2 \\
\quad \opTok{FUNC} & lexical & allus. & function-word swap & \textit{ac} / \textit{et} & 28 & 5.6 & 9.6 & 2.2 \\
\quad \opTok{META} & lexical & allus. & figurative or metonymic & \textit{oculos} / \textit{lumina} & 17 & 3.4 & 1.7 & 4.7 \\
\quad \opTok{CONSTR} & lexical & allus. & same idea, other construction & \textit{tulerunt} / \textit{sustineret} & 11 & 2.2 & 2.6 & 1.8 \\
\quad \opTok{NE-SUB} & lexical & allus. & other proper name & \textit{Thebas} / \textit{Colchos} & 9 & 1.8 & 2.2 & 1.5 \\
\quad \opTok{HYPER} & lexical & allus. & more general word & \textit{quercus} / \textit{arbor} & 9 & 1.8 & 0.9 & 2.6 \\
\quad \opTok{ANT} & lexical & allus. & opposite & \textit{tenebrae} / \textit{lux} & 9 & 1.8 & 1.3 & 2.2 \\
\quad \opTok{PRON} & lexical & allus. & pronoun for noun, or reverse & \textit{Caesar} / \textit{ille} & 7 & 1.4 & 1.7 & 1.1 \\
\quad other (4) & lexical & allus. & \opTok{NUM}, \opTok{HYPO}, \opTok{CO-HYPO}, \opTok{SYN-DIST} & n/a & 9 & 1.8 & 1.3 & 2.2 \\
\quad none named & lexical & allus. & no nameable relation & n/a & 158 & 31.4 & 37.1 & 26.6 \\
\midrule
\multicolumn{9}{@{}l}{\emph{Unlinked words (share of reuse words; \opTok{DEL}, \opTok{OUT}: of source words)}} \\
\opTok{INS} & residual & no match & reuse word added inside the stretch & n/a & 2,480 & 6.5 & 4.8 & 8.7 \\
\opTok{NEW} & residual & no match & reuse word outside the stretch & n/a & 28,646 & 74.9 & 68.7 & 83.0 \\
\opTok{MARK} & structural & marker & citing formula & \textit{ut ait Maro} & 872 & 2.3 & 3.8 & 0.3 \\
\opTok{DEL} & residual & no match & source word dropped inside the stretch & n/a & 2,824 & 7.3 & 6.0 & 8.9 \\
\opTok{OUT} & residual & no match & source word outside the stretch & n/a & 29,700 & 76.6 & 71.5 & 83.2 \\
\midrule
\multicolumn{9}{@{}l}{\emph{Patterns read off the links}} \\
\opTok{QUOTE} & pattern & n/a & $\geq$4 in-order copies (of pairs) & n/a & 324 & 21.7 & 37.2 & 0.8 \\
\opTok{REORDER} & pattern & n/a & link out of source order (of links) & n/a & 533 & 8.5 & 6.4 & 16.2 \\
\opTok{DISPERSE} & pattern & n/a & links cross clauses (of pairs) & n/a & 71 & 4.8 & 4.8 & 4.7 \\}
\def\DataAllOpsCoreRows{%
\multicolumn{9}{@{}l}{\emph{Links (share of links)}} \\
\opTok{COPY} & form & verb. & same form, spelling folded & \textit{frigidus} / \textit{frigidus} & 4,828 & 77.2 & 85.2 & 48.0 \\
\opTok{INFLECT} & form & verb. & same lemma, other form & \textit{regem} / \textit{reges} & 870 & 13.9 & 9.3 & 30.8 \\
\opTok{SUBST} & lexical & allus. & other lemma, same slot & \textit{pugnam} / \textit{calamitatem} & 503 & 8.0 & 4.7 & 20.5 \\
\opTok{SPLIT} & cardinality & allus. & two reuse words, one source & n/a & 40 & 0.6 & 0.7 & 0.4 \\
\opTok{MERGE} & cardinality & allus. & one reuse word, two sources & n/a & 9 & 0.1 & 0.1 & 0.2 \\
\midrule
\multicolumn{9}{@{}l}{\emph{Unlinked words (share of reuse words; \opTok{DEL}, \opTok{OUT}: of source words)}} \\
\opTok{INS} & residual & no match & reuse word added inside the stretch & n/a & 2,480 & 6.5 & 4.8 & 8.7 \\
\opTok{NEW} & residual & no match & reuse word outside the stretch & n/a & 28,646 & 74.9 & 68.7 & 83.0 \\
\opTok{MARK} & structural & marker & citing formula & \textit{ut ait Maro} & 872 & 2.3 & 3.8 & 0.3 \\
\opTok{DEL} & residual & no match & source word dropped inside the stretch & n/a & 2,824 & 7.3 & 6.0 & 8.9 \\
\opTok{OUT} & residual & no match & source word outside the stretch & n/a & 29,700 & 76.6 & 71.5 & 83.2 \\
\midrule
\multicolumn{9}{@{}l}{\emph{Patterns read off the links}} \\
\opTok{QUOTE} & pattern & n/a & $\geq$4 in-order copies (of pairs) & n/a & 324 & 21.7 & 37.2 & 0.8 \\
\opTok{REORDER} & pattern & n/a & link out of source order (of links) & n/a & 533 & 8.5 & 6.4 & 16.2 \\
\opTok{DISPERSE} & pattern & n/a & links cross clauses (of pairs) & n/a & 71 & 4.8 & 4.8 & 4.7 \\}

\usepackage{acl}
\usepackage{eso-pic} %

\newcommand{\releaselinks}{%
  Code: \href{https://github.com/julianschelb/retexo}{https://github.com/julianschelb/retexo}\\
  Data: \href{https://hf.co/datasets/julian-schelb/latin-classical-intertextuality-edit-scripts}{https://hf.co/datasets/julian-schelb/latin-classical-intertextuality-edit-scripts}\\
  Demo: \href{https://julianschelb.github.io/retexo/demo/}{https://julianschelb.github.io/retexo/demo/}}

\usepackage{times}
\usepackage{latexsym}
\usepackage{graphicx}
\usepackage{amsmath}
\usepackage{amssymb}
\usepackage{booktabs}
\usepackage{array}
\usepackage{tabularx} %
\usepackage{colortbl} %
\usepackage{pifont}
\usepackage{multirow} %
\usepackage{enumitem}
\usepackage{listings}
\usepackage[most]{tcolorbox}
\usepackage{subcaption}
\usepackage{tikz}
\usetikzlibrary{arrows.meta, positioning, fit, backgrounds, shapes.geometric, calc}
\usepackage{pgfplots}
\pgfplotsset{compat=1.17}

\definecolor{userpromptbg}{HTML}{eff6fe} %
\definecolor{userpromptborder}{HTML}{6c8ebf}
\definecolor{userprompttitle}{HTML}{aec7e8}
\definecolor{userprompttitletext}{HTML}{2c3e50}

\newtcolorbox{examplebox}[1][]{
    listing only,
    enhanced,
    colback=userpromptbg,
    colframe=userpromptborder,
    colbacktitle=userprompttitle,
    coltitle=userprompttitletext,
    arc=2mm,
    boxrule=1pt,
    left=2mm,
    right=2mm,
    top=2mm,
    bottom=2mm,
    fonttitle=\bfseries\small,
    title=#1,
    toptitle=0.5mm,
    bottomtitle=0.5mm,
    listing options={
        basicstyle=\ttfamily\tiny,
        breaklines=true,
        columns=flexible,
        keepspaces=true,
    },
    drop shadow,
}

\definecolor{accent}{HTML}{1F4E79}   %
\definecolor{accent2}{HTML}{2B6CB0}  %
\definecolor{citgreen}{HTML}{2E7D32} %
\definecolor{cfred}{HTML}{C62828}    %
\definecolor{figmut}{HTML}{6B7280}   %
\definecolor{fbgreen}{HTML}{1B7F3B}  %
\definecolor{shadepos}{HTML}{2B6CB0} %
\definecolor{shadeneg}{HTML}{D9822B} %

\newcommand{\opTok}[1]{{\small\ttfamily #1}}
\newcommand{\stk}[1]{\begin{tabular}[b]{@{}c@{}}#1\end{tabular}}
\newcommand{\stkl}[1]{\begin{tabular}[c]{@{}l@{}}#1\end{tabular}}
\tikzset{
  fbox/.style   = {draw=accent, rounded corners=2pt, line width=0.7pt, fill=accent!4,
                   align=center, inner sep=4pt, font=\scriptsize},
  fdata/.style  = {draw=figmut, rounded corners=2pt, line width=0.5pt, fill=black!2,
                   align=center, inner sep=4pt, font=\scriptsize},
  fflow/.style  = {-{Stealth[length=1.8mm]}, line width=0.7pt, accent},
  flbl/.style   = {font=\tiny\itshape, color=figmut},
}

\usepackage[T1]{fontenc}

\usepackage[utf8]{inputenc}

\usepackage{microtype}

\title{Learning Word-Level Explanations of Text Reuse \\ in Latin Literature with Little Annotation}

\author{Julian Schelb$^\dagger$ \and Michael Wittweiler$^\diamond$ \and Marie Revellio$^\ddagger$ \\
  $^\dagger$Department of Computer and Information Science, University of Konstanz \\
  $^\ddagger$Department of Latin Philology, University of Konstanz \\
  $^\diamond$Institute of Archaeology, Classical Philology and Ancient Studies, University of Zurich \\
  \texttt{\{firstname.lastname\}@uni-konstanz.de} \\
  \texttt{\{firstname.lastname\}@uzh.ch}}

\defcitealias{DBLP:conf/emnlp/StahlbergK20}{Stahlberg et al.}
\begin{document}
\AddToShipoutPictureFG*{\AtPageUpperLeft{\put(\LenToUnit{0.5\paperwidth},\LenToUnit{-1.3cm}){\makebox[0pt][c]{\small\textit{Preprint. October 2026.}}}}}
\maketitle

\begin{abstract}
Detecting an intertextual reference is only the first step; philologists also need to know \emph{how} a later author reworked the source.
We explain a reuse pair word by word: each word of the reusing passage is first aligned to its source word, then labeled with the operation that transformed it, such as a copy, a change of inflection, or a substitution.
Such word-level operations are expensive to annotate, and only a few corpora annotated at this level exist for text reuse in Latin literature.
We describe a procedure that needs only a limited number of annotations: synthetic pairs, built by applying known operations to Latin text, train a first model, and active learning chooses the pairs an annotator corrects.
Among several architectures, a model that links and labels together performs best: a pointer network that, for every reuse word, chooses a source word and names the operation in the same step.
We show that with active learning, 200 pairs chosen by uncertainty are enough for the model to match one trained on every annotated pair of the 1,490 references of the Loci Similes benchmark.
We release our code, word-level predictions for all references, and an interactive demo.\footnote{\releaselinks}

\end{abstract}

\section{Introduction}
\label{sec:intro}
Identifying intertextual references is a central task in classical philology: it shows how later works engage with the texts they inherit, what inspired them, and which works their authors knew.
Recent work on Latin literature has established sentence-pair baselines that frame the problem as a three-way classification of a (query, candidate) pair into \textit{no reference}, \textit{verbatim reference} (\textit{cit.}), or \textit{allusion} (\textit{cf.})~\citep{locisimiles2026}.
The two reference types follow the convention of critical editions and quotation indices~\citep{revellio2022zitate, schropp2024goldstandard}, but they divide reuse only coarsely, and scholars draw the line between them differently.
More importantly, such a label indicates only \emph{whether} two passages are related; it does not explain \emph{how} the later author reworked the source.

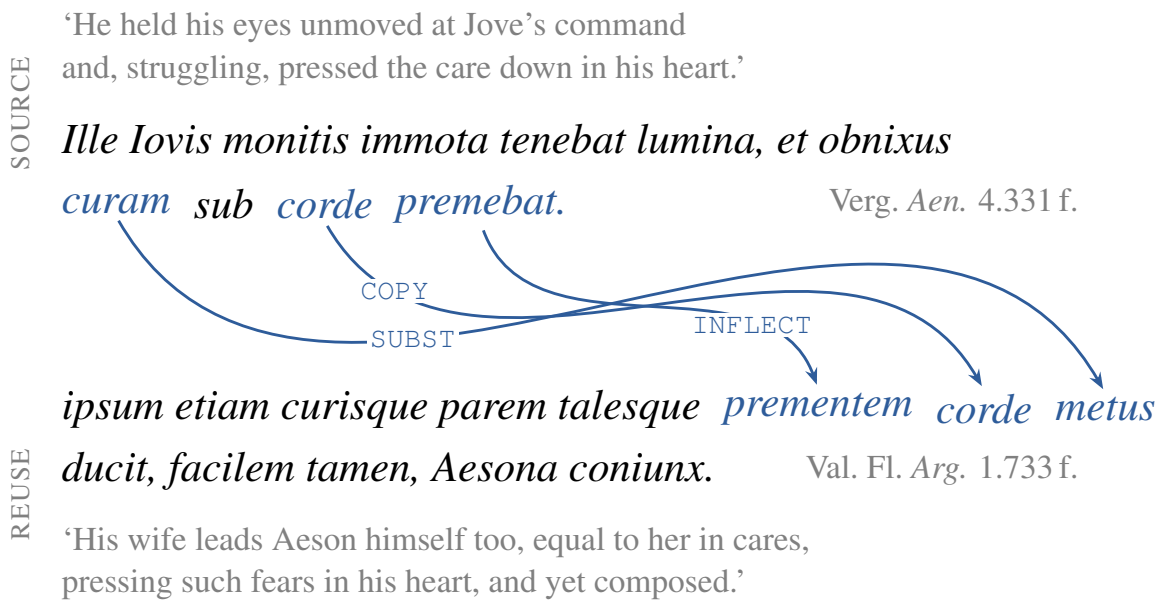
\begin{figure}[t]
\centering
\definecolor{linkblue}{HTML}{2E5C9A}
\resizebox{\columnwidth}{!}{%
\begin{tikzpicture}[x=1pt, y=1pt,
  lat/.style={inner sep=0.5pt, font=\footnotesize\itshape},
  lword/.style={inner sep=0.5pt, font=\footnotesize\itshape, text=linkblue},
  en/.style={inner sep=0.5pt, font=\scriptsize, text=gray},
  cite/.style={font=\scriptsize, text=gray},
  tag/.style={font=\tiny\ttfamily, text=linkblue, fill=white, inner sep=0.6pt},
  link/.style={linkblue, line width=0.6pt, -{Stealth[length=3.5pt]}, shorten >=1pt, shorten <=1pt}]
  \node[cite, rotate=90, anchor=center] at (-8,82) {\textsc{source}};
  \node[cite, rotate=90, anchor=center] at (-8,2) {\textsc{reuse}};
  \node[en, anchor=west] at (0,100) {`He held his eyes unmoved at Jove's command};
  \node[en, anchor=west] at (0,91) {and, struggling, pressed the care down in his heart.'};
  \node[lat, anchor=west] (sl1) at (0,76) {Ille Iovis monitis immota tenebat lumina, et obnixus};
  \node[lword, anchor=west] (s_curam) at (0,63) {curam};
  \node[lat, right=3.5pt of s_curam] (s_sub) {sub};
  \node[lword, right=3.5pt of s_sub] (s_corde) {corde};
  \node[lword, right=3.5pt of s_corde] (s_premebat) {premebat.};
  \node[cite, anchor=east] at (218,63) {Verg.\ \textit{Aen.} 4.331\,f.};
  \node[lat, anchor=west] (rl1) at (0,20) {ipsum etiam curisque parem talesque};
  \node[lword, right=3.5pt of rl1] (q_prementem) {prementem};
  \node[lword, right=3.5pt of q_prementem] (q_corde) {corde};
  \node[lword, right=3.5pt of q_corde] (q_metus) {metus};
  \node[lat, anchor=west] (rl2) at (0,7) {ducit, facilem tamen, Aesona coniunx.};
  \node[cite, anchor=east] at (218,7) {Val.\ Fl.\ \textit{Arg.} 1.733\,f.};
  \node[en, anchor=west] at (0,-8) {`His wife leads Aeson himself too, equal to her in cares,};
  \node[en, anchor=west] at (0,-17) {pressing such fears in his heart, and yet composed.'};
  \draw[link] (s_curam.south) to[out=-60,in=120] node[tag, pos=0.33]{SUBST} (q_metus.north);
  \draw[link] (s_corde.south) to[out=-60,in=120] node[tag, pos=0.14]{COPY} (q_corde.north);
  \draw[link] (s_premebat.south) to[out=-70,in=110] node[tag, pos=0.75]{INFLECT} (q_prementem.north);
\end{tikzpicture}}
\caption{\textbf{Example of reworked reuse.} Valerius Flaccus reworks half a line of Virgil: \textit{corde} is kept (\opTok{COPY}), \textit{premebat} returns as a participle (\opTok{INFLECT}), and \textit{metus} stands in the slot of \textit{curam} (\opTok{SUBST}).}
\label{fig:intro_script}
\vspace{-0.5cm}
\end{figure}

Figure~\ref{fig:intro_script} illustrates this with Virgil and its reuse in Valerius Flaccus.
In Virgil, Aeneas presses his care down in his heart; Valerius Flaccus reuses the expression for Aeson, who holds down his fears.
He keeps \textit{corde} (heart), the word that makes the reference recognizable; he turns the verb \textit{premebat} (he pressed) into the participle \textit{prementem} (pressing) to fit his own sentence; and he replaces the object \textit{curam} (care) with \textit{metus} (fears), a word of similar meaning.
Together, the three operations show that the expression of an emotion pressed down in the heart survives intact, while its grammar and one of its words are adapted: Valerius Flaccus alludes to Virgil rather than quoting him verbatim.
Such an explanation makes the boundary between \textit{cit.} and \textit{cf.} measurable, allows comparison of how different authors handle their sources, and shows at a glance which words were copied and which were changed.

We therefore explain a reuse pair at the level at which philologists describe it: the word.
Each word of the reusing passage is linked to the source word it corresponds to, or to none, and each link is labeled with the operation that relates the two words: a copy, a change of inflection, or a substitution.
We call the result an \emph{edit script}. For a verbatim reference, a script shows which words were copied and whether their order changed; for an allusion, it shows which words survive and which were replaced.
Producing such a script resembles word alignment between a source and a reuse passage, with two differences. First, each link carries a label that names the operation the later author applied. Second, unlike alignment in translation or other multilingual settings, most words of a reusing passage have \emph{no} counterpart in the source, so deciding that a word has no source is most of the task.

Training a model to produce such scripts requires labeled data. For Latin, such data exists only for 100 Bible reuses~\citep{moritz-etal-2016-non}.
Extending it by hand is slow, since an annotator must decide a link and an operation for every word of both passages.
We therefore combine two ways to reduce the number of annotations required.
The first is synthetic data. In the spirit of \citet{DBLP:journals/nature/TrinhWLHL24}, we build pairs by applying known operations to clean Latin text, so their scripts are correct by construction. A model trained on these pairs learns what a script looks like before any pair is annotated.
The second is active learning~\citep{settles-craven-2008-analysis}. The model pre-annotates the corpus and selects the pairs it is least sure of. An annotator corrects these pairs, and the model trains again.
Together, the two turn annotation into correction, starting from zero annotated pairs.
We simulate this procedure on the 1,490 references of the Loci Similes benchmark~\citep{locisimiles2026} and find that with 200 corrected pairs, the model matches one trained on every annotated pair.

To produce both the link and the operation for each word, we use a pointer network~\citep{DBLP:conf/nips/VinyalsFJ15} and compare it with eleven other systems. Finally, we analyze whether the predicted scripts also serve as evidence of reuse. We contribute:

\begin{itemize}[leftmargin=10pt, itemsep=0.5pt,topsep=3pt]
    \item \textbf{Pointer model for edit scripts} that links each reuse word to a source word, or to none, and labels the link in the same step, making reuse detection explainable word by word.
    \item \textbf{Annotation procedure} that combines synthetic pairs with uncertainty-based active learning to reduce labeling effort.
    \item \textbf{Analysis of predicted operations}, examining whether they carry evidence of reuse and if they distinguish verbatim references from allusions.
\end{itemize}

\section{Related Work}
\label{sec:relwork}

Our work builds on four lines of research: the detection of intertextual references in Latin, monolingual word alignment, the identification of edit operations, and active learning from a cold start.

\begin{table*}[t]
\centering
\small
\setlength{\tabcolsep}{4pt}%
\resizebox{\textwidth}{!}{%
\begin{tabular}{@{}lllllrrrr@{}}
\toprule
\textbf{Label} & \textbf{Signal} & \textbf{Association} & \textbf{Meaning} & \textbf{Example} & \multicolumn{4}{c}{\textbf{Annotation}} \\
\cmidrule(lr){6-9}
 & & & & (source / reuse) & \textbf{$n$} & \textbf{\%} & \textbf{\textit{cit.} \%} & \textbf{\textit{cf.} \%} \\
\midrule
\DataAllOpsCoreRows
\bottomrule
\end{tabular}}
\caption{\textbf{Label inventory and distribution.} Labels are grouped by the signal that decides them and by their association with the reference (Association: verbatim, allusion, marker, or no match). $n$ is the count in the annotation and \% its share of the operation. \opTok{INS}, \opTok{NEW}, \opTok{DEL}, \opTok{OUT}, and the patterns are derived from the links and not annotated.}
\label{tab:new_labels}
\label{tab:new_allops}
\end{table*}

\subsection{Intertextuality Detection in Latin}
Computational work on Latin intertextuality has framed detection as discovery since \citet{bamman2008logic}: a system surfaces candidate parallels, and a reader judges them~\citep{duan2025quantitative}.
Tesserae~\citep{coffee2012intertextuality, coffee2013tesserae} matches word pairs shared by two texts, and \citet{scheirer2016sense} extended it to shared meaning through latent semantic indexing.
\citet{revellio2022zitate} and \citet{DBLP:journals/dhq/SchroppKRF24} detected short quotations with rule-based and n-gram matching.
Such methods capture verbatim and near-verbatim reuse well, but miss related passages with little shared vocabulary.

Embeddings narrow that gap.
\citet{DBLP:conf/naacl/BurnsBLCD21} ranked intertextual phrases with static word vectors trained on lemmatized text, and \citet{DBLP:conf/latech/ManjavacasLK19} treated allusion detection as retrieval with fastText embeddings.
\citet{gong_augmented_2025} replaced static vectors with contextual LatinBERT embeddings~\citep{DBLP:journals/corr/abs-2009-10053}, and \citet{DBLP:conf/ranlp/Riemenschneider23} introduced sentence-transformer models for ancient languages.

\citet{locisimiles2026} compared these models on 1{,}490 expert-verified references labeled as verbatim reference or allusion; we take our corpus, folds, and pair-level baselines from this benchmark.
Beyond Latin, \citet{smith2014detecting} and \citet{liebl-burghardt-2020-shakespeare} detected reuse at corpus scale, where passages are aligned only as a by-product and the links carry no labels.
All of these approaches locate a parallel; none describes how the source was transformed, which we do word by word.

\subsection{Monolingual Word Alignment}

Describing reuse word by word is a word alignment problem within a single language, but most alignment methods were developed for translation and predict links without labels.
Statistical aligners learn word-translation probabilities from many sentence pairs and introduced a null word for words without a counterpart~\citep{brown-etal-1993-mathematics, ostling2016efficient}.
\citet{sultan-etal-2014-back} aligned sentences of the same language with rules, linking words that are similar in meaning and whose contexts are linked as well.
\citet{jalili-sabet-etal-2020-simalign} aligned words by the similarity of their contextual embeddings without parallel data, and \citet{dou-neubig-2021-word} fine-tuned such embeddings on parallel text.
\citet{zenkel-etal-2020-end} read alignments from an alignment layer trained inside a neural translation model.
\citet{nagata-etal-2020-supervised} cast alignment as span prediction with a multilingual encoder fine-tuned on a few hundred hand-aligned sentences.

Two methods produce links and labels in one step.
\citet{DBLP:conf/emnlp/StahlbergK20} predicted span-level edits together with a tag for each edit.
Pointer networks~\citep{DBLP:conf/nips/VinyalsFJ15} select a position of the input as their output.%
For Latin, word alignment has been studied between a text and its translation~\citep{yousef-etal-2022-automatic-translation}.
These methods were not designed for reuse, where most words of a passage have no counterpart and every link carries a label.

\subsection{Identifying Edit Operations}
Paraphrase typologies catalog the operations by which one text becomes another.
\citet{bhagat-hovy-2013-squibs} listed 25 types of quasi-paraphrase, \citet{vila2014paraphrase} organized paraphrase types on four linguistic levels, and \citet{wahle-etal-2023-paraphrase} showed that language models find the type of a paraphrase much harder to detect than the paraphrase itself.
Our label set is such a typology, narrowed to literary reuse and attached to links between words.
The closest precedent is \citet{moritz-etal-2016-non}, who classified each word-level difference between a Bible passage and its reuse in Ancient Greek and Latin, with operations defined by lexical resources.
They reported corpus-level statistics but did not ask whether such scripts can be \emph{learned}.%

Token-level editing models predict an operation for each word of a single text.
\citet{DBLP:conf/emnlp/MalmiKRMS19} tagged words to be kept or deleted and inserted phrases from a fixed vocabulary, and \citet{DBLP:conf/emnlp/MallinsonSMG20} separated tagging from a pointer that reorders words.
We adopt the pointer~\citep{DBLP:conf/nips/VinyalsFJ15} to choose a source word, but apply it across two passages and let a word remain unlinked.

\subsection{Cold-Start Active Learning}
Active learning selects the examples from which a model learns most, but it needs a model that already knows something about the task.
\citet{yuan-etal-2020-cold} showed that uncertainty sampling fails from a cold start, when the model has not yet learned the task.
\citet{DBLP:journals/nature/TrinhWLHL24} avoided human labels altogether by training on synthetic examples that are correct by construction.
We adapt this idea: our synthetic pairs teach the task before any pair is annotated, which removes the cold start.

\citet{baldridge-osborne-2004-active} found that pre-annotation by the model lowers annotation cost, also in combination with instance selection.
\citet{ambati2010active} applied this to word alignment, where annotators corrected the links a model was least sure of.
We combine both: annotators correct the model's scripts on the pairs it selects.
Each round restarts from the synthetic model, since \citet{ash2020warm} showed that warm-started networks generalize worse.

\section{Method}

\begin{figure*}[t]
\centering
\resizebox{\textwidth}{!}{%
\begin{tikzpicture}[x=1pt, y=1pt,
  box/.style={draw=accent, rounded corners=2pt, line width=0.7pt, fill=accent!4, inner sep=4pt, align=center, font=\footnotesize},
  data/.style={draw=figmut, rounded corners=2pt, line width=0.5pt, fill=black!2, inner sep=4pt, align=center, font=\footnotesize},
  stage/.style={font=\scriptsize\bfseries, text=accent, inner sep=0pt},
  note/.style={font=\scriptsize, text=figmut, inner sep=0pt, align=left},
  rowl/.style={font=\scriptsize\itshape, inner sep=0pt, anchor=east},
  coll/.style={font=\scriptsize\itshape, inner sep=0pt, anchor=west, rotate=50},
  lword/.style={inner sep=0.5pt, font=\scriptsize\itshape, text=accent2},
  gword/.style={inner sep=0.5pt, font=\scriptsize\itshape, text=figmut},
  ltag/.style={font=\tiny\ttfamily, text=accent, inner sep=0pt},
  ntag/.style={font=\tiny\ttfamily, text=figmut, inner sep=0pt},
  link/.style={accent2, line width=0.5pt, -{Stealth[length=2.5pt]}, shorten >=1pt, shorten <=1pt},
  flow/.style={-{Stealth[length=1.8mm]}, line width=0.7pt, accent},
  thin/.style={-{Stealth[length=1.4mm]}, line width=0.5pt, figmut}]
\node[stage] at (0,96) {Input};
\node[data, anchor=north west, text width=112pt] (in) at (0,90) {%
  {\scriptsize\color{figmut}source $S$}\\ \textit{curam sub corde premebat}\\[2pt]
  {\scriptsize\color{figmut}reuse $Q$}\\ \textit{prementem corde metus}};
\node[note, anchor=north, align=center] at ([yshift=-3pt]in.south) {read as \texttt{[CLS]}\,$S$\,\texttt{[SEP]}\,$Q$\,\texttt{[SEP]},\\ once in each order};
\node[stage, anchor=north west] at (0,18) {Encoder};
\node[box, anchor=north west, text width=112pt] (enc) at (0,10) {LatinBERT\\[1pt] {\scriptsize one vector per word, $\mathbf{h}_s$ and $\mathbf{h}_t$}};
\node[stage, anchor=north west, align=left] at (128,104) {Scores $\ell(t,s)$\\ per reuse word};
\node[rowl] at (181,76.5) {prementem};
\draw[fill=accent!8, draw=black!25, line width=0.3pt] (184,71) rectangle (199,82);
\draw[fill=accent!3, draw=black!25, line width=0.3pt] (199,71) rectangle (214,82);
\draw[fill=accent!7, draw=black!25, line width=0.3pt] (214,71) rectangle (229,82);
\draw[fill=accent!60, draw=accent, line width=0.9pt] (229,71) rectangle (244,82);
\draw[fill=accent!15, draw=black!25, line width=0.3pt] (244,71) rectangle (259,82);
\draw[fill=accent!1, draw=black!25, line width=0.3pt] (259,71) rectangle (274,82);
\node[rowl] at (181,65.5) {corde};
\draw[fill=accent!6, draw=black!25, line width=0.3pt] (184,60) rectangle (199,71);
\draw[fill=accent!3, draw=black!25, line width=0.3pt] (199,60) rectangle (214,71);
\draw[fill=accent!63, draw=accent, line width=0.9pt] (214,60) rectangle (229,71);
\draw[fill=accent!7, draw=black!25, line width=0.3pt] (229,60) rectangle (244,71);
\draw[fill=accent!8, draw=black!25, line width=0.3pt] (244,60) rectangle (259,71);
\draw[fill=accent!1, draw=black!25, line width=0.3pt] (259,60) rectangle (274,71);
\node[rowl] at (181,54.5) {metus};
\draw[fill=accent!38, draw=accent, line width=0.9pt] (184,49) rectangle (199,60);
\draw[fill=accent!4, draw=black!25, line width=0.3pt] (199,49) rectangle (214,60);
\draw[fill=accent!7, draw=black!25, line width=0.3pt] (214,49) rectangle (229,60);
\draw[fill=accent!10, draw=black!25, line width=0.3pt] (229,49) rectangle (244,60);
\draw[fill=accent!27, draw=black!25, line width=0.3pt] (244,49) rectangle (259,60);
\draw[fill=accent!2, draw=black!25, line width=0.3pt] (259,49) rectangle (274,60);
\node[coll] at (187,84) {curam};
\node[coll] at (202,84) {sub};
\node[coll] at (217,84) {corde};
\node[coll] at (232,84) {premebat};
\node[coll] at (247,84) {$\varnothing$ \opTok{INS/NEW}};
\node[coll] at (262,84) {$\varnothing$ \opTok{MARK}};
\node[font=\tiny, text=accent, inner sep=0pt] at (191.5,44) {$\checkmark$};
\node[font=\tiny\ttfamily\bfseries, text=accent, inner sep=0pt] at (206.5,44) {DEL};
\node[font=\tiny, text=accent, inner sep=0pt] at (221.5,44) {$\checkmark$};
\node[font=\tiny, text=accent, inner sep=0pt] at (236.5,44) {$\checkmark$};
\node[note, anchor=north west] at (152,36) {a source word that no row\\ chooses carries the implicit\\ label \opTok{DEL} or \opTok{OUT}};
\node[note, anchor=north west, text width=104pt] at (152,4) {score: dot product $f(\mathbf{h}_t)^{\top} g(\mathbf{h}_s)$ $+$ pair MLP $(\mathbf{x}_{t,s})$ $+$ $\mathbf{w}^{\top}\boldsymbol{\phi}(t,s)$, with $\boldsymbol{\phi}$ the 23 lexical features};
\node[stage] at (320,96) {Heads};
\node[box, anchor=north west, text width=104pt] (link) at (320,90) {Link head\\ {\scriptsize softmax over the row\\ $\to a_t$: the source word, or none}};
\node[box, anchor=north west, text width=104pt] (op) at (320,42) {Operation head\\ {\scriptsize softmax at the chosen word\\ $\to y_t$: what was done to it}};
\node[box, anchor=north west, text width=104pt] (stretch) at (320,-10) {Stretch head\\ {\scriptsize $p(q_t$ not reused$)$, added to the \opTok{INS/NEW} null}};
\draw[thin] (stretch.west) to[out=180, in=-90] (251.5,48);
\node[stage] at (482,96) {Decoding};
\node[box, anchor=north west, text width=82pt, align=left] (dec) at (482,90) {{\scriptsize \textbullet\ average both orders\\ \textbullet\ link if score $> \theta$\\ \textbullet\ one-to-one assignment\\ \textbullet\ restore splits\\ \textbullet\ \opTok{COPY} gate}};
\node[stage, anchor=north west] at (482,32) {Output};
\node[note, anchor=north west] (otitle) at (482,18) {edit script: a typed alignment};
\node[lword, anchor=east] (s0) at (518,4) {curam};
\node[gword, anchor=east] (s1) at (518,-6) {sub};
\node[lword, anchor=east] (s2) at (518,-16) {corde};
\node[lword, anchor=east] (s3) at (518,-26) {premebat};
\node[ntag, anchor=east] at ($(s1.west)+(-2,0)$) {DEL};
\node[lword, anchor=west] (q0) at (542,4) {prementem};
\node[ltag, anchor=west] (t0) at ($(q0.east)+(2,0)$) {INFLECT};
\node[lword, anchor=west] (q1) at (542,-6) {corde};
\node[ltag, anchor=west] (t1) at ($(q1.east)+(2,0)$) {COPY};
\node[lword, anchor=west] (q2) at (542,-16) {metus};
\node[ltag, anchor=west] (t2) at ($(q2.east)+(2,0)$) {SUBST};
\draw[link] (s0.east) to[out=0, in=180] (q2.west);
\draw[link] (s2.east) to[out=0, in=180] (q1.west);
\draw[link] (s3.east) to[out=0, in=180] (q0.west);
\begin{scope}[on background layer]
\node[data, fit=(otitle)(s0)(s3)(q0)(q2)(t0)(t1)(t2), inner sep=5pt] (out) {};
\end{scope}
\draw[flow] (in.west) to[out=200, in=160] (enc.west);
\draw[flow] (enc.east) to[out=0, in=180] (150,65.5);
\draw[flow] (276,71) to (link.west);
\draw[flow] (276,60) to (op.west);
\draw[flow] (link.east) to (dec.west |- link.east);
\draw[flow] (op.east) to[out=0, in=180] ([yshift=5pt]dec.south west);
\draw[flow] (dec.south) to (out.north);
\end{tikzpicture}}
\caption{\textbf{Architecture of our stretch-and-link pointer.} For each reuse word, the model scores every source word and two null positions, which link to no source word (\opTok{INS}/\opTok{NEW} or \opTok{MARK}). Three heads decide: the link head chooses a source word or a null position, the operation head names what was done to the chosen word (such as \opTok{COPY} or \opTok{SUBST}), and the stretch head estimates whether the reuse word is reused at all and adds this estimate to the \opTok{INS}/\opTok{NEW} null. A source word that no reuse word chooses (here \textit{sub}) is labeled \opTok{DEL}.}
\label{fig:architecture}
\end{figure*}
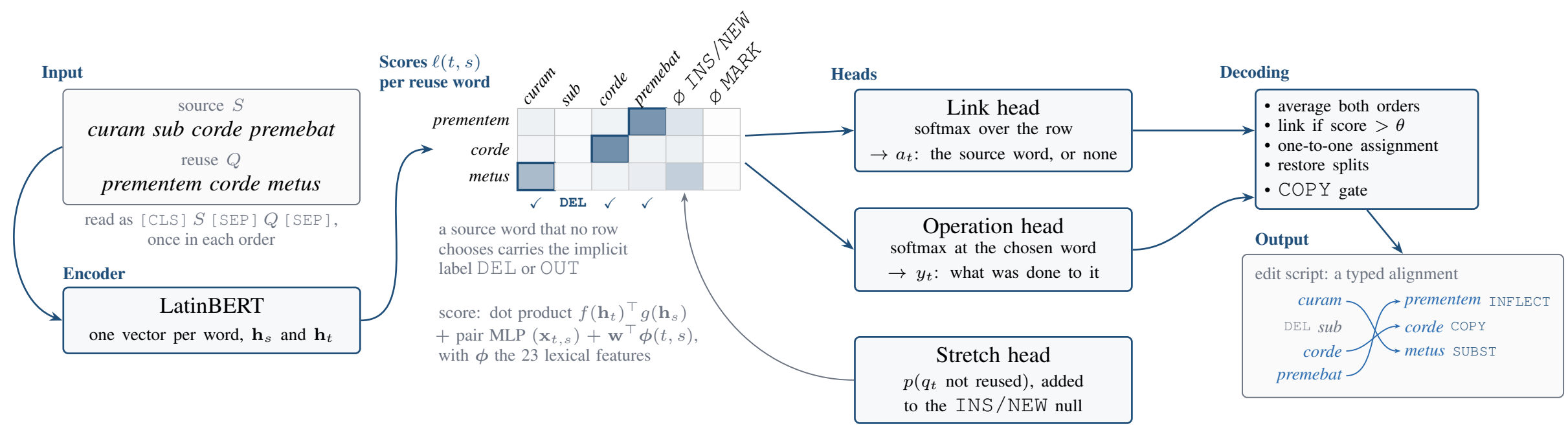
\label{sec:new_method}

We model the explanation of a reuse pair as the prediction of an edit script: a model that links each reuse word to its source and names what was done to it in one step, synthetic pairs that teach the model the task before any pair is annotated, and an active-learning loop in which annotators correct the few pairs the model chooses.

\subsection{Task Definition}
\label{subsec:new_task}

An edit script describes, word by word, how a reuse relates to its source.
Formally, a pair consists of two sequences of words, a source $S = s_1 \dots s_n$ and a reuse $Q = q_1 \dots q_m$.
An edit script assigns every reuse word $q_t$ a link $a_t \in \{1, \dots, n\} \cup \{\varnothing\}$, the position of its source word or $\varnothing$ if it has none, and a label $y_t$ that names the operation.
Links are one-to-one, with one exception described below.
A source word that no link claims also receives a label, so that a script accounts for every word of both passages.
In Figure~\ref{fig:intro_script}, three reuse words are linked, each with a different label, and the source word \textit{sub} is claimed by no link.

\subsection{Label Set}
\label{subsec:new_labels}

The labels follow the paraphrase typology of \citet{vila2014paraphrase} and the lexical relations of historical text reuse~\citep{moritz-etal-2016-non}; Table~\ref{tab:new_labels} lists them with examples. They fall into two kinds: labels of linked words, which have a counterpart in the other passage, and labels of unlinked words, which have none. For analysis purposes, labels are also grouped by \emph{signal}, the kind of evidence that decides them, and words by \emph{association} (verbatim, allusion, marker, or no match).
Finally, \emph{patterns} combine multiple operations.
They are derived from a finished script and never annotated (Appendix~\ref{app:annotation}).

\paragraph{Linked words.}
A linked word is a copy (\opTok{COPY}), an inflection (\opTok{INFLECT}), or a substitution (\opTok{SUBST}) of its source word, or one half of a word that was split or merged (\opTok{SPLIT}, \opTok{MERGE}).

\paragraph{Unlinked words.}
An unlinked reuse word is an insertion (\opTok{INS}) inside a stretch of reused words and new text (\opTok{NEW}) outside it; a word of a citing formula such as \textit{ut ait Maro} (as Maro says) is labeled \opTok{MARK}.
An unlinked source word is likewise a deletion (\opTok{DEL}) inside the stretch, as \textit{sub} in the figure, and text not taken up (\opTok{OUT}) outside it.

\subsection{Model}
\label{subsec:new_model}

The model is a pointer network~\citep{DBLP:conf/nips/VinyalsFJ15}: for every reuse word, it chooses a source word, or none, and names the operation at the word it chose (Figure~\ref{fig:architecture}).
We call it the \emph{stretch-and-link pointer}, after the two decisions it couples: whether a word lies inside a reused stretch, and which source word it links to.

\paragraph{Encoder.}
The encoder turns every word of the pair into a vector.
LatinBERT~\citep{DBLP:journals/corr/abs-2009-10053} reads the pair as one sequence, \texttt{[CLS]}~$S$~\texttt{[SEP]}~$Q$~\texttt{[SEP]}, so that every reuse word attends to every source word, and a second time with the passages swapped.
The vector of a word is the mean of its subword states, $\mathbf{h}_s$ for a source word and $\mathbf{h}_t$ for a reuse word.

\paragraph{Link head.}
The link head decides which source word a reuse word links to (if any).
For each reuse word, it scores every source word and two null positions, one for a word without a source and one for a word of a citing formula, and takes one softmax over them.
A link score is the sum of three terms: a scaled dot product of the two word vectors, a small network over the joint vector of the pair, and a weighted sum of 23 lexical features looked up in the Latin resources, such as a shared lemma or a WordNet relation (more details in Appendix~\ref{app:model}).
The dot product favors words that \emph{look} alike in context.
The other two terms let the model link words that \emph{mean} alike but differ in form, which is the case for substitutions.

\paragraph{Operation head.}
The operation head names the operation of a link, such as \opTok{INFLECT} or \opTok{SUBST}.
For the source word the link head identified, it takes a second softmax over the same joint vector and lexical features to predict the operation.
The probability of a labeled link is the product of the probabilities of its link and its label.
Linking and labeling share the encoder and are trained together, but each has its own softmax.
A separate binary decision marks an unlinked word as part of a citing formula.

\paragraph{Stretch head.}
The stretch head complements the link head by deciding, from a word's neighbors, whether it belongs to a reused stretch (verbatim, allusion, marker) or not (no match). A word inside a stretch usually has a source, other words usually do not. The probability of no match is added, with a learned weight, to the score of the no-source null position.

\paragraph{Training and decoding.}
All heads are trained together, each with a cross-entropy loss against the annotated script.
For every reuse word, the loss is the sum of four terms: (1.) the link loss, which compares the link head's distribution with the annotated source word or null position; (2.) the operation loss, which is computed at the annotated source word and only for linked words; (3.) the marker loss of the binary citing-formula decision; and (4.) the association loss of the stretch head.
The loss of a pair is the mean over its reuse words.
At inference, the scores of both reading directions are averaged, a reuse word is linked only if its best score exceeds a threshold $\theta$ set on the dev fold, and the links are made one-to-one (Appendix~\ref{app:setup_details}).
One rule overrides the operation head: two words that are identical after spelling and enclitics are normalized form a copy.

\subsection{Active Learning}
\label{subsec:new_method_al}

A synthetic pair is built from clean Latin text by applying known operations to it, so its script is correct by construction~\citep{DBLP:conf/emnlp/AwasthiSGGP19, DBLP:journals/nature/TrinhWLHL24}.
It teaches the \emph{shape} of the task, that most words have no source and how a link is drawn, but not the \emph{conventions} of a human annotation, such as which spelling variants count as copies; those must come from annotated pairs, and the question is how few.

The loop starts from the model trained on synthetic pairs alone.
In each round the model ranks an unannotated pool, annotators correct the scripts of the top-ranked pairs, and the model is trained again from the synthetic model on every pair corrected so far~\citep{ash2020warm}.
We compare four rankings: (1) random selection; (2) uncertainty sampling~\citep{settles-craven-2008-analysis, shen2018deep}, which ranks a pair by the share of its words whose link decision has a probability below 0.8; (3) hard-operation density, which prefers pairs with many predicted inflections and substitutions; and (4) expected corrections, which combines the two by counting the words that are uncertain or carry a hard operation.

\section{Data}
\label{sec:data_new}

\begin{table*}[t]
\centering
\scriptsize
\setlength{\tabcolsep}{1.5pt}
\resizebox{\textwidth}{!}{%
\begin{tabular}{@{}llcccccccc@{}}
\toprule
 & & \multicolumn{3}{c}{\textbf{Links}} & \multicolumn{4}{c}{\textbf{Labels}} & \\
\cmidrule(lr){3-5}\cmidrule(lr){6-9}
\textbf{System} & \textbf{Method} & \textbf{P} & \textbf{R} & \textbf{F1} & \textbf{macro F1} & \textbf{INFLECT} & \textbf{SUBST} & \textbf{MARK} & \textbf{Invented} \\
\midrule
\multicolumn{10}{@{}l}{\textit{Links, then the rule}} \\
Statistical aligner & \citealp{brown-etal-1993-mathematics} & \rv{ac.em_aligner.p} & \rv{ac.em_aligner.r} & \rv{ac.em_aligner.f} & \rv{ac.em_aligner.l3} & \rv{ac.em_aligner.infl} & \rv{ac.em_aligner.subst} & \rv{ac.em_aligner.frame} & \rv{ac.em_aligner.inv} \\
Rule-based aligner & \citealp{sultan-etal-2014-back} & \rv{ac.sultan.p} & \rv{ac.sultan.r} & \rv{ac.sultan.f} & \rv{ac.sultan.l3} & \rv{ac.sultan.infl} & \rv{ac.sultan.subst} & \rv{ac.sultan.frame} & \rv{ac.sultan.inv} \\
Embedding similarity & \citealp{jalili-sabet-etal-2020-simalign} & \rv{ac.sim_aligner.p} & \rv{ac.sim_aligner.r} & \rv{ac.sim_aligner.f} & \rv{ac.sim_aligner.l3} & \rv{ac.sim_aligner.infl} & \rv{ac.sim_aligner.subst} & \rv{ac.sim_aligner.frame} & \rv{ac.sim_aligner.inv} \\
Embedding similarity (FT) & \citealp{dou-neubig-2021-word} & \rv{ac.sim_aligner_ft.p} & \rv{ac.sim_aligner_ft.r} & \rv{ac.sim_aligner_ft.f} & \rv{ac.sim_aligner_ft.l3} & \rv{ac.sim_aligner_ft.infl} & \rv{ac.sim_aligner_ft.subst} & \rv{ac.sim_aligner_ft.frame} & \rv{ac.sim_aligner_ft.inv} \\
Translation-model aligner & \citealp{zenkel-etal-2020-end} & \rv{ac.nmt_aligner.p} & \rv{ac.nmt_aligner.r} & \rv{ac.nmt_aligner.f} & \rv{ac.nmt_aligner.l3} & \rv{ac.nmt_aligner.infl} & \rv{ac.nmt_aligner.subst} & \rv{ac.nmt_aligner.frame} & \rv{ac.nmt_aligner.inv} \\
Span aligner (SpanAlign) & \citealp{nagata-etal-2020-supervised} & \rv{ac.span_aligner.p} & \rv{ac.span_aligner.r} & \rv{ac.span_aligner.f} & \rv{ac.span_aligner.l3} & \rv{ac.span_aligner.infl} & \rv{ac.span_aligner.subst} & \rv{ac.span_aligner.frame} & \rv{ac.span_aligner.inv} \\
\midrule
\multicolumn{10}{@{}l}{\textit{Links and labels together}} \\
Span-pair scorer (Seq2Edits) & \citetalias{DBLP:conf/emnlp/StahlbergK20}, 2020 & \rv{ac.span_pair.p} & \rv{ac.span_pair.r} & \rv{ac.span_pair.f} & \rv{ac.span_pair.l3} & \rv{ac.span_pair.infl} & \rv{ac.span_pair.subst} & \rv{ac.span_pair.frame} & \rv{ac.span_pair.inv} \\
Pointer, plain (pointer net.) & \citealp{DBLP:conf/nips/VinyalsFJ15} & \rv{ac.plain_pointer.p} & \rv{ac.plain_pointer.r} & \rv{ac.plain_pointer.f} & \rv{ac.plain_pointer.l3} & \rv{ac.plain_pointer.infl} & \rv{ac.plain_pointer.subst} & \rv{ac.plain_pointer.frame} & \rv{ac.plain_pointer.inv} \\
\midrule
\multicolumn{10}{@{}l}{\textit{Ours}} \\
Stretch-and-link pointer & this work & \rv{ac.ours.p} & \rv{ac.ours.r} & \rv{ac.ours.f} & \rv{ac.ours.l3} & \rv{ac.ours.infl} & \rv{ac.ours.subst} & \rv{ac.ours.frame} & \rv{ac.ours.inv} \\
Ours + span aligner & this work & \rv{ac.ours_plus_span_aligner.p} & \rv{ac.ours_plus_span_aligner.r} & \rv{ac.ours_plus_span_aligner.f} & \rv{ac.ours_plus_span_aligner.l3} & \rv{ac.ours_plus_span_aligner.infl} & \rv{ac.ours_plus_span_aligner.subst} & \rv{ac.ours_plus_span_aligner.frame} & \rv{ac.ours_plus_span_aligner.inv} \\
\midrule
\multicolumn{10}{@{}l}{\textit{Reference}} \\
Copy floor (identical forms) & n/a & \rv{ac.copy_floor.p} & \rv{ac.copy_floor.r} & \rv{ac.copy_floor.f} & \rv{ac.copy_floor.l3} & \rv{ac.copy_floor.infl} & \rv{ac.copy_floor.subst} & \rv{ac.copy_floor.frame} & \rv{ac.copy_floor.inv} \\
Annotated links + rule typer & n/a & \rv{ac.gold_links.p} & \rv{ac.gold_links.r} & \rv{ac.gold_links.f} & \rv{ac.gold_links.l3} & \rv{ac.gold_links.infl} & \rv{ac.gold_links.subst} & \rv{ac.gold_links.frame} & \rv{ac.gold_links.inv} \\
\bottomrule
\end{tabular}}
\caption{\textbf{Performance comparison of alignment architectures.} Links are scored by precision (P), recall (R), and F1 under sure/possible scoring, operation labels by F1, and citing formulas by \opTok{MARK} span F1. Invented counts links per pair on random re-pairings, which contain no reuse. Method gives the work each system is based on; FT marks the fine-tuned embedding aligner. All other values are $\times100$ and averaged over five folds.}
\label{tab:res_ac}
\vspace{-0.4cm}
\end{table*}

We use two kinds of data: annotated references for evaluation and correction, and synthetic pairs for the first stage of training.

\subsection{Annotations}
\label{subsec:new_annotation}

We annotate the 1,490 references of the Loci Similes benchmark~\citep{locisimiles2026}, 858 verbatim references (\textit{cit.}) and 632 allusions (\textit{cf.}), and keep its pair labels and folds.
Each annotation records the links, their labels, the spans of citing formulas, and whether a link is sure or only possible, as 655 of the 6,250 links are.
The annotation was produced with the correction loop of Section~\ref{subsec:new_method_al} (guideline and statistics in Appendix~\ref{app:annotation}).

\subsection{Synthetic Training Pairs}
\label{subsec:new_synthetic}

A synthetic pair is built from a seed passage, which serves as the source, and an unrelated host passage.
Multiple words of the seed are copied, re-inflected, substituted, split, or merged, and set into the host passage in shuffled order (Appendix~\ref{app:setup_details}).
The synthetic pairs contain far more substitutions than the annotated ones, and most replace a word with an unrelated one and teach a model \emph{where} a substitution happens better than \emph{what} counts as one.

\section{Experimental Setup}
\label{sec:setup_new}

We describe the compared systems and their training, then the experiments and the metrics.

\subsection{Systems}
\label{subsec:new_systems}

We compare our model with eight established methods, which fall into two groups by what they predict (Table~\ref{tab:res_ac}).
The first group consists of six aligners that predict only the links.
A fixed rule then labels each link: identical forms are a copy, a shared lemma is an inflection, and anything else is a substitution.
The second group, a span-pair scorer and a plain pointer network, predicts link and label together, as our model does.
We also report a variant of our model that averages its link scores with those of the span aligner.
Two reference systems help to interpret the scores.
The copy floor links identical words and nothing else, which any trained system should exceed.
The second applies the fixed rule to the annotated links, which separates a system's link errors from its label errors.

\subsection{Training}
\label{subsec:new_training}
\label{subsec:new_splits}

We use the five folds of the Loci Similes benchmark.
For test fold $k$, fold $(k+1) \bmod 5$ serves as the dev fold, on which the inference settings of each system are set.
The remaining three folds are used for training, with 20\,\% of their pairs held out as a validation sample for early stopping.
All trained systems use the same data and schedule, so that the way a system produces a script is the only difference between them.
The training set of a fold consists of 25,000 synthetic pairs, the annotated training pairs, and as many pairs without reuse.
A system is trained for one epoch on the synthetic pairs and then on the annotated pairs until the macro F1 on the validation sample stops improving.
Every system that outputs link scores is decoded like our model (Section~\ref{subsec:new_model}), with the threshold $\theta$ set per fold on the dev fold.
Further details, including the learning-rate search and the hyperparameters, are given in Appendix~\ref{app:setup_details}.

\subsection{Experiments}
\label{subsec:new_exps}
\label{subsec:new_ac}
\label{subsec:new_svr}
\label{subsec:new_al}
\label{subsec:new_downstream}

We run three experiments: a comparison of architectures, a comparison of active-learning strategies, and a downstream classification task.

\paragraph{Architectures and ablations.}
The first experiment determines which system produces the most accurate scripts. Every system runs on the five folds and is scored on each fold's annotated test pairs and on random re-pairings of its passages. The random pairs contain no reuse, so that the links a system invents can be counted.
We conduct two ablations of our model.
The first adds its components one at a time, so that the difference to the previous step, is the contribution of that component.
The second labels the same links in three ways: by the fixed rule, by a classifier trained afterwards, and by our model, which labels while it links.

\paragraph{Active-learning strategies.}
The second experiment determines how few annotations the model needs.
We first measure a learning curve: our model is trained on subsets of the annotated pairs, once after the synthetic stage and once without it.
We then simulate the annotation process of Section~\ref{subsec:new_annotation} with each of the four strategies of Section~\ref{subsec:new_method_al}: the annotation of a pair is revealed only when the loop selects it, and the cost is reported in corrected pairs and corrected words.
Each strategy is compared with random selection inside the loop and not with the learning curve, because only the random loop shares its starting model and training schedule, so that the choice of pairs is the only difference.
The budgets and two control runs are given in Appendix~\ref{app:setup_details}.

\paragraph{Downstream classification.}
The third experiment examines whether the predicted scripts carry evidence of reuse.
The first task separates the true reference from the other 99 candidates the retriever of the benchmark proposes for each reusing passage; the second separates verbatim references from allusions.
For both, a logistic regression uses only counts derived from a decoded script, with no word overlap and no lexicon, and is compared with the similarity score of the benchmark (Appendix~\ref{app:setup_details}).

\subsection{Metrics}
\label{subsec:new_metrics}

A predicted script is compared with the annotated one link by link, using precision, recall, and F1.
Link F1 counts a link as correct when it connects the right two words, with precision against sure and possible links and recall against sure links only~\citep{och-ney-2003-systematic}. Macro F1, our main metric, also requires the label to be correct~\citep{bryant-etal-2017-automatic} and is the unweighted mean over the five labels.
Invented links are counted on random re-pairings, where every link is wrong; the downstream tasks keep the metrics of the benchmark~\citep{locisimiles2026} (Appendix~\ref{app:metrics}).

\section{Results}
\label{sec:results_new}
\label{sec:results}

We report the results of the three experiments in turn, followed by an error analysis.
Unless stated otherwise, all values are means over the five test folds, multiplied by 100 (metrics in Section~\ref{subsec:new_metrics}).

\subsection{Architecture Comparison}
\label{subsec:res_ac}

Table~\ref{tab:res_ac} scores every system of Section~\ref{subsec:new_systems}: the links in the first block of columns, the labels in the second, and the links a system invents on pairs without reuse in the last.
Our model reaches the best macro F1 and the best F1 on both hard labels, inflection and substitution.
Every trained system clears the copy floor, but the systems differ little on links and much on labels, nowhere more than on substitution.
The span aligner finds links as well as ours and invents the fewest, but the rule that names its links loses the substitutions a learned label recovers.
Averaging our link scores with the span aligner's gives the best links but, named by the rule, a lower macro F1.
The remaining gap to the annotated links named by the same rule is therefore a gap in the hard links, which Section~\ref{subsec:res_errors} takes apart.

Among our components, the ablations attribute the gain to the lexical evidence and the copy rule at decoding, while the stretch head is positive on three folds only (Table~\ref{tab:res_ablations}, Appendix~\ref{app:additional}).
On the same links, a classifier trained afterwards labels worse than the shared rule, and our model, which labels while it links, beats the rule: the operation head is part of the pointer, not an addition to it.

\begin{figure}[t]
\centering
\includegraphics[width=\columnwidth]{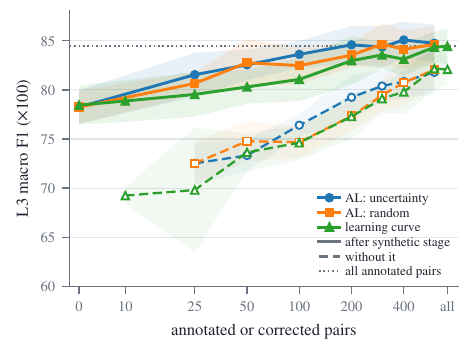}
\caption{\textbf{Performance vs.\ annotation effort trade-off.} Macro F1 against the number of corrected pairs; the learning curve uses random annotated pairs. Mean over five folds, bands $\pm1$ standard deviation.}
\label{fig:res_al}
\vspace{-0.5cm}
\end{figure}

\subsection{Learning from Few Annotations}
\label{subsec:res_little}

With the synthetic stage, a few hundred annotated pairs yield a higher macro F1 than all annotated pairs yield without it (Figure~\ref{fig:res_al}).
The gain is largest where annotations are fewest.
The synthetic stage therefore replaces annotation, most of all the first few hundred pairs.
Starting from the synthetic model alone, uncertainty sampling matches the model trained on all annotated pairs after 200 corrected pairs (\rv{al.A.uncertainty.200} against \rv{al.all.syn}).
At this budget it also exceeds random selection (\rv{al.A.random.200}).
The two other strategies do not exceed random selection, so what helps is the uncertainty of the model about a pair, not the predicted difficulty of its labels.
The result holds when the threshold is fixed, and the pairs selected by uncertainty are hardly longer than random ones.
Without the synthetic stage it reverses, as expected of a model that has not yet learned the task~\citep{yuan-etal-2020-cold}.
Since every selected pair is corrected and not annotated from scratch, the cost per pair is also lower than these counts suggest.

\subsection{Operation Profiles as Evidence of Reuse}
\label{subsec:res_downstream}

\begin{figure}[t]
\centering
\includegraphics[width=\columnwidth]{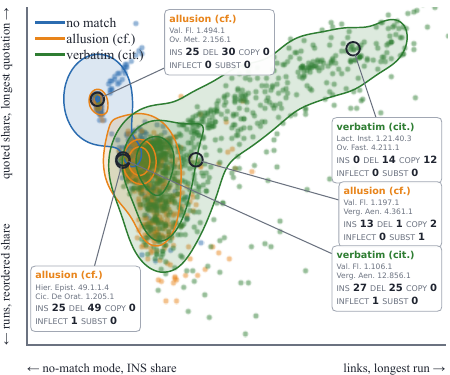}
\caption{\textbf{Projection of operation-derived features} for verbatim references, allusions, and no-match pairs. Allusions overlap with both verbatim references and no-match pairs, while verbatim references and no-match pairs are distinct. Contours show density.}
\label{fig:ana_projection}
\vspace{-0.5cm}
\end{figure}

Features derived from our scripts exceed the similarity score of the benchmark on per-query macro F1 (\rv{ds.ours.f1} against \rv{ds.similarity.f1}) and on Recall@10 (\rv{ds.ours.r10} against \rv{ds.similarity.r10}), while passage length alone stays near chance.
Systems with better scripts in Table~\ref{tab:res_ac} also provide better evidence (Appendix~\ref{app:downstream}).
What separates a reference from a no-match pair is how many words the model links and how confidently: a no-match pair receives few links, most of them with low probability.
Our scripts separate verbatim references from allusions nearly as well as the annotated scripts (macro F1 \rv{ds.ours.type} against \rv{ds.gold.type}).
A verbatim reference reuses more words, in longer runs, and with fewer changes.
Substitutions and reorderings are more frequent in allusions (Table~\ref{tab:new_labels}), but they separate the two types only in combination with these features.
In the projection of Figure~\ref{fig:ana_projection}, allusions lie between verbatim references and no-match pairs.
This explains why a classifier that must assign all three classes at once confuses allusions with no-match pairs.

\subsection{Error Analysis}
\label{subsec:res_errors}

The main source of error is the decision whether a word has a source at all, not the label of a link once it is drawn.
We analyze the errors of our model on the two hard labels on fold 4, so the counts are indicative (Tables~\ref{tab:res_errors} and~\ref{tab:res_bounds}, Appendix~\ref{app:errors}).

Over the five test folds, the predicted label distribution matches the annotated one within a percentage point for every label.
The errors therefore concern where a label is placed, not how often it is predicted.
Most missed inflections and substitutions are words the model left without a link.
Most wrongly predicted ones are enclitics and spelling variants, which the annotation counts as copies but the synthetic pairs never show as such; the copy rule at decoding addresses these.
Most of the rest are invented links.
A manual inspection of the predictions confirms the two kinds of error.
The model invents links between unrelated words, such as \textit{crurum} and \textit{boum} (Virgil, \textit{Georgics} 3.51, and Jerome, \textit{Epistula} 52.1.2).
It also misses links between related words, such as \textit{miseros nautas} and \textit{miseris nautis} (Statius, \textit{Thebaid} 3.316, and Valerius Flaccus 1.567).

\section{Discussion and Conclusion}
\label{sec:discussion}
\label{sec:conclusion}

We presented a model that explains text reuse word by word, and test how much annotation it requires.
First, annotation can start without annotated pairs: after a synthetic stage, 200 pairs that the model selected for correction match a model trained on all annotated pairs (Section~\ref{subsec:res_little}).
Second, linking and labeling are best learned in one model, which exceeds every combination of an aligner and a rule on inflection and substitution (Section~\ref{subsec:res_ac}).
Third, the predicted edit scripts carry evidence of reuse: features derived from them identify true references better than similarity scores and separate verbatim references from allusions nearly as well as the annotated scripts (Section~\ref{subsec:res_downstream}).
A predicted script gives philologists a first analysis of a reference: it shows which words a author reused and how each was changed.
In future work, a generator that substitutes related words instead of unrelated ones could teach the model what counts as a substitution, not only where one occurs.
We release the code, the predicted scripts of all 1,490 references, and a review app in which every script can be inspected word by word.

\section*{Limitations}
\label{sec:limitations}

\paragraph{Annotation Robustness.}
The corrections of the annotation loop were made by a single annotator, so the annotation reflects one reading of the guideline and no inter-annotator agreement could be measured; where the guideline leaves room, as in naming the relation of a substitution or marking a link as possible, another annotator may decide differently.

\paragraph{Coverage of the Lexical Resources.}
A relation that the Latin resources do not record cannot be named by the rule and leaves the lexical evidence silent, and the resources are silent on most of the substitutions our model misses (Appendix~\ref{app:resources}).
Patterns defined over clause boundaries inherit the punctuation of modern editions.

\section*{Acknowledgements}
This work was supported by the Deutsche Forschungsgemeinschaft (DFG, German Research Foundation), project no.\ 382880410, \textit{Quoting as a narrative strategy: A digital-hermeneutic analysis of intertextual references in Christian Late Antiquity}.

\section*{AI Usage Statement}

Language model-based AI tools (Claude and GitHub Copilot) were used as coding assistants
during implementation. Additionally, AI-based tools were used for grammar and spell
checking.

\bibliography{
    bibliography/anthology,
    bibliography/intertextuality,
    bibliography/alignment,
    bibliography/edit_operations,
    bibliography/active_learning,
    bibliography/resources,
    bibliography/benchmark,
    bibliography/methods,
    bibliography/related_work}
\bibliographystyle{acl_natbib}

\appendix

\section{Resources and Normalization}
\label{app:resources}

\paragraph{Resources.} The generator, the rule typer, and the model's lexical evidence draw on the same resources: the lemmatizer, the Collatinus decliner, and the proper-name list of the Classical Language Toolkit~\citep{johnson2021cltk} (CLTK 1.5.0 with its Latin models of February 2022); the Latin WordNet~\citep{franzini2019evaluation, santoro-etal-2025-exploring} through its Exeter API,\footnote{\url{https://latinwordnet.exeter.ac.uk}} queried in September 2026 and cached (3,524 lemma entries, released with the code); and the lemmatized word2vec vectors released with \citet{DBLP:conf/naacl/BurnsBLCD21} (CBOW, 300 dimensions, window 5, minimum count 100, 125,256 lemmas; read with gensim 4.4.0).
The parts of speech and morphological features shown to the annotators, and the parses of the rule-based aligner, come from the Latin models of Stanza~\citep{qi-etal-2020-stanza} (1.11.0, ITTB package).
They are consulted in a fixed order (identity, morphology, WordNet relation, distributional neighbor, named entity), so the rule and the evidence are deterministic for these versions.

\paragraph{Normalization.} Forms are compared after lowercasing, u/v and i/j regularization, and the stripping of diacritics, punctuation, and attached enclitics; the copy rule at decoding and the scorer use this normalized form, so a spelling variant counts as a copy, as the annotation requires.

\paragraph{Coverage.} A substitution that no resource records keeps its label in the annotation, but the rule cannot name its relation and the evidence is silent on it; this holds for most of the substitutions our model misses (Section~\ref{subsec:res_errors}).
Patterns over clause boundaries (\opTok{DISPERSE}) inherit the punctuation of modern editions and so measure the editor's segmentation as well as the author's practice; the pattern stays because rule-based quotation filters use clause breaks in the same form~\citep{DBLP:journals/dhq/SchroppKRF24}.

\section{Setup Details}
\label{app:setup_details}

\paragraph{Model equations.}
\label{app:model}
The link head of Section~\ref{subsec:new_model} scores every source word $s$ for a reuse word $t$ from their vectors $\mathbf{h}_s$ and $\mathbf{h}_t$, and takes one softmax over the source words and the two null positions:
\begin{equation}
\ell(t,s) = \frac{f(\mathbf{h}_t)^{\top} g(\mathbf{h}_s)}{\sqrt{d}} \negthinspace+\negthinspace \operatorname{MLP}_{\text{loc}}(\mathbf{x}_{t,s}) \negthinspace+\negthinspace \mathbf{w}^{\negthinspace\top}\negthinspace\boldsymbol{\phi}(t,s),
\label{eq:new_locate}
\end{equation}
where $f$ and $g$ are linear projections of width $d$, $\mathbf{x}_{t,s} = [\mathbf{h}_t; \mathbf{h}_s; |\mathbf{h}_t - \mathbf{h}_s|; \mathbf{h}_t \odot \mathbf{h}_s]$ is the pair's joint vector, and $\boldsymbol{\phi}(t,s)$ holds the lexical features.
At the chosen source word the operation head names the operation with a second softmax,
\begin{equation}
p(a_t = s, y_t = k) = p(a_t = s) \cdot \operatorname{softmax}_k(\mathbf{z}_{t,s}),
\label{eq:new_name}
\end{equation}
where $\mathbf{z}_{t,s} = \operatorname{MLP}_{\text{name}}(\mathbf{x}_{t,s}) + \mathbf{V}\boldsymbol{\phi}(t, s)$ holds one score per label.
The 23 features of $\boldsymbol{\phi}$ record whether the two words share a form or a lemma, whether WordNet relates them, whether both are names, whether an enclitic hides a shared stem, how far apart they stand, and whether each resource knows the word at all (Appendix~\ref{app:resources}); they enter the link score and the label score, so a shared lemma helps the model find an inflected word and not only name it.

\paragraph{Training.} Our model is trained with AdamW, the encoder at a learning rate of $2\times10^{-5}$ and the heads at $10^{-3}$, in batches of 32 pairs of at most 256 subword tokens, with seed 1: one epoch over the 25,000 synthetic pairs in both reading directions, then up to 16 passes over the annotated training pairs (669 to 747 per fold), each mixed with 600 fresh synthetic pairs and one pair without reuse per annotated pair, until the macro F1 on the validation sample has not improved for two epochs (the best epoch is kept).
The link loss weighs unlinked words less and the label loss rare labels more; a substitution without a named relation is trained as the set of lexical labels.
Every other trained system follows the same schedule, with the same search over three learning rates (half the default, the default, double) on 5,000 synthetic pairs and the dev fold of test fold~0; the translation-model aligner stops on its translation loss, since its alignment layer never sees a link.

\paragraph{Compute.} The encoder is LatinBERT with \rv{compute.params.encoder}\,M parameters; the heads of our model add \rv{compute.params.heads}\,M.
All runs were made on one server with two NVIDIA A40 GPUs (46\,GB each), each run on one GPU; the runs behind the tables and figures took about \rv{compute.hours.reported} GPU hours, the learning-rate search a further \rv{compute.hours.tuning}.

\paragraph{Pairs without reuse.} Each pairs a benchmark passage with a random passage of a cited author, with the passage that shares the most words with it, or with the passage next to its true source; without them every script on a no-match pair would look like evidence of reuse.

\paragraph{Decoding.} The scores of both reading directions are averaged, a reuse word is linked if its best score exceeds $\theta$, the links are made one-to-one by an optimal assignment~\citep{kuhn1955hungarian}, and a source word claimed by two reuse words is restored as a split.
$\theta$ is the value with the highest word accuracy on the dev fold; tuning it by macro F1 instead changes only the two embedding-similarity rows (fine-tuned 0.733 instead of 0.675, frozen 0.683 instead of 0.665), at the price of more invented links.

\paragraph{Generator.} A seed passage gives up zero to four fragments of one to twenty words (one fragment in 62\,\% of pairs, two in 31\,\%); related substitutes (synonyms, hypernyms, derivational relatives, proper names) come from the resources of Appendix~\ref{app:resources}, unrelated ones from the masked language model LaBerta~\citep{DBLP:conf/acl/Riemenschneider23}.\footnote{\url{https://huggingface.co/bowphs/LaBerta}}
Fragments are shuffled within and between fragments at a rate of 0.3 each, one pair in ten receives a citing formula, and the pool is sampled so that every rare label reaches a minimum count; each fold's training set also holds about 1,000 held-out synthetic pairs.

\paragraph{Splits.} Source passages may recur across folds, as in the benchmark (149 of the 1,239 source segments do).
The validation sample is 20\,\% of the training pairs, grouped by reusing passage (168 to 189 pairs per fold).
No passage of the test fold, the dev fold, or the validation sample appears in a synthetic pair, a host passage, or a pair without reuse; the two aligners that learn from text alone read the dev and test passages as unlabeled text.

\paragraph{Systems.} The statistical and the translation-model aligner learn from the passages as unlabeled text and never see a link; the rule-based aligner, the copy floor, and the merged row train nothing.
The span-pair scorer's candidate spans are pruned with the scores of our model without its label head, and the merged row averages the link scores of our model and the span aligner before decoding.

\paragraph{Active learning.} The learning curve trains our model on 100, 300, or 600 annotated pairs, or on all of them, nested and drawn in one fixed order per fold; the model trained on synthetic pairs alone is its point at zero.
The loop's pool is the fold's training pairs outside the validation sample, with nested budgets of 25, 50, 100, 200, and 400 pairs after a first batch of 10.
The main run tunes $\theta$ on the dev fold, which uses annotated pairs the loop never acquired; a second run therefore fixes $\theta$ at 0.8, and a third removes the synthetic stage.
Uncertainty takes the share, not the count, of a pair's uncertain words, so long pairs are not chosen for their length~\citep{shen2018deep}; without the synthetic stage every strategy's first batch is the same random draw.

\paragraph{Downstream classification.} The classifier sees the 30 counts and shares of Table~\ref{tab:ana_features}, read off the decoded script, and no word, lemma, or lexicon.
It is trained on four folds at ten no-match candidates per positive and scored per query on the fifth, as in the benchmark~\citep{locisimiles2026}, with its threshold set on the training rows by the benchmark's plateau rule.

\paragraph{Metrics.}
\label{app:metrics}
Links and labels are scored as typed edits~\citep{bryant-etal-2017-automatic}.
Link F1 counts a predicted link as correct when it connects the right reuse word to the right source word, whatever its label, with precision against sure and possible links and recall against sure links only~\citep{och-ney-2003-systematic}; we report F1 rather than the alignment error rate, which tracks downstream quality worse~\citep{fraser-marcu-2007-squibs}.
Per-label F1 counts a label only at the correct source word, and citing formulas are scored as exact spans.
Invented links, counted per pair on random re-pairings where every link is wrong, and link recall measure the two ways a system can fail, inventing a link and declining one.
Macro F1, the headline number, requires link and label to be correct together and averages five labels without weighting: \opTok{COPY}, \opTok{INFLECT}, \opTok{SUBST} (with \opTok{SPLIT} and \opTok{MERGE}), \opTok{INS} (with \opTok{NEW} and \opTok{MARK}), and \opTok{DEL} (with \opTok{OUT}).
The downstream tasks keep the benchmark's scores~\citep{locisimiles2026}: macro F1 averaged over query segments and Recall@10 for finding the reference, and macro F1 over the true references for verbatim reference against allusion.

\section{Annotation}
\label{app:annotation}

\begin{figure*}[p]
\centering
\includegraphics[width=\textwidth]{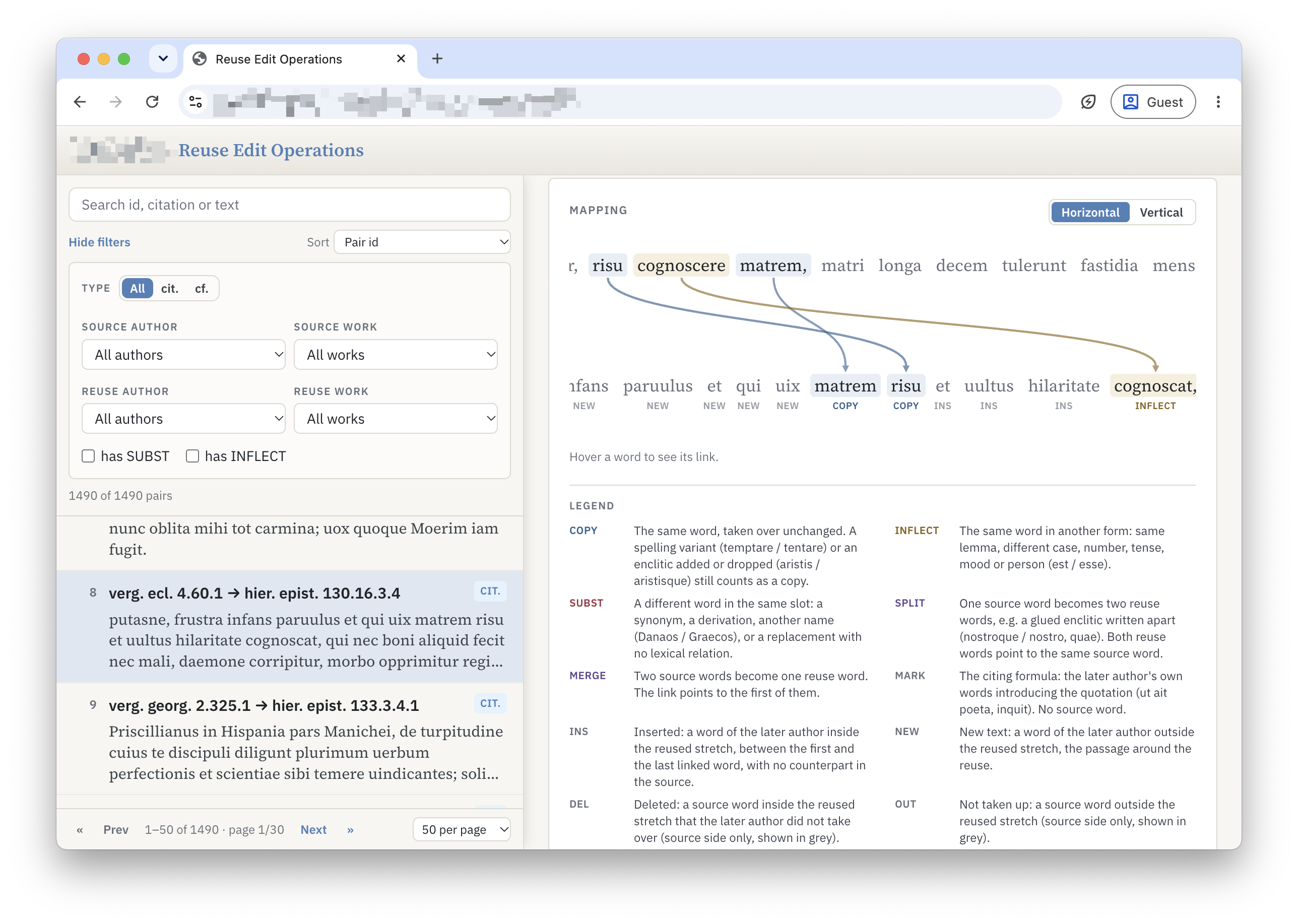}
\caption{\textbf{The review app.} The predicted script of one reference (Virgil, \textit{Eclogues} 4.60, in Jerome, \textit{Epistula} 130.16): the list of pairs with filters on the left, the links as arrows from the source words (top) to the reuse words (bottom) with their operations, and the legend. The app is released with the code.}
\label{fig:demo}
\end{figure*}

\paragraph{Provenance.} The 1,490 references are those of the benchmark~\citep{locisimiles2026}, 945 of them from the Valerius Flaccus intertext database~\citep{DBLP:conf/naacl/BurnsBLCD21, dexter2024database}.
Their word-level annotation was produced in rounds of the correction loop of Section~\ref{subsec:new_method_al}: for each pair the loop chose, the annotators saw both passages, each word with its dictionary form, part of speech, and related words from a lemmatizer and the Latin WordNet, and the model's current script, and corrected the script, one line per reuse word naming its source word or none and the operation, and one line marking the reused stretch in both passages.
The rounds continued until every pair had a corrected script; twenty-three pairs repeat the text of another pair and count once in the downstream experiment.
The released review app (Figure~\ref{fig:demo}) shows the predicted script of every pair in the same form: the links drawn as arrows between the two passages, each reuse word with its operation, and the words without a link marked as new text, insertions, deletions, or text not taken up.

\paragraph{Guideline.} A word is taken from the source only inside a reused stretch, a run in which at least two content words reproduce the source close together and in the same order, or one rare and unmistakable phrase; shared function words, forms of \textit{esse}, and common verbs are never evidence on their own.
Inside a stretch the annotation is complete: the reuse word that stands where a source word stood is linked to it even when the two are unrelated (\opTok{SUBST}), and a word that only changed its ending is \opTok{INFLECT}.
The words with which the reusing author introduces or attributes the quotation form the citing formula (\opTok{MARK}).
For a substitution the annotator names the closest relation of Table~\ref{tab:new_labels_full} that can be defended from the two passages, or none, and marks a link as possible instead of sure where the evidence is thin.

\paragraph{Conventions.} Spelling is folded before forms are compared (case, u/v, i/j, ae/e, an attached \textit{-que}), so a word that differs only there is a \opTok{COPY}; where this line runs is a convention that editors may draw differently.
The lemma decides between inflection and substitution, so a derivational relative is a substitution.
A split is the only case of a source word linked twice, almost always an enclitic split off its word.
\opTok{INS}, \opTok{NEW}, \opTok{DEL}, \opTok{OUT}, and the patterns are derived from the links: the reused stretch of a passage runs from its first to its last linked word, and a reordering marks each reuse word whose link lies outside the longest chain of links in source order.
An unrelated substitution can carry the meaning of the reuse: Jerome (\textit{Epistula} 77.11.2) replaces \textit{Herculeas} in Virgil's \textit{laudes Herculeas} (\textit{Aeneid} 8.287f.) with \textit{femineas}, and the praise of Hercules becomes the praise of women.

\paragraph{Corpus statistics.} Both passages of a pair average 26 words, but a pair has only 4.2 links, and 89.5\,\% of the links are sure; Table~\ref{tab:new_labels} gives every label's share, and Table~\ref{tab:new_labels_full} adds the relations.
Most reuse words are new text and most source words text not taken up, so the decision that a word has \emph{no} source is the most frequent one a model makes, and the one that most often hides a missed link; copies dominate the links, which is why the metric averages over labels and not over links.
Allusions substitute synonyms more often than verbatim references, which more often swap function words, so a model needs to recognize synonyms most where the passages share the fewest words.

\begin{table*}[t]
\centering
\small
\setlength{\tabcolsep}{4pt}%
\resizebox{\textwidth}{!}{%
\begin{tabular}{@{}lllllrrrr@{}}
\toprule
\textbf{Label} & \textbf{Signal} & \textbf{Association} & \textbf{Meaning} & \textbf{Example} & \multicolumn{4}{c}{\textbf{Annotation}} \\
\cmidrule(lr){6-9}
 & & & & (source / reuse) & \textbf{$n$} & \textbf{\%} & \textbf{\textit{cit.} \%} & \textbf{\textit{cf.} \%} \\
\midrule
\DataAllOpsSlimRows
\bottomrule
\end{tabular}}
\caption{\textbf{Label inventory with the relations of \opTok{SUBST}.} As Table~\ref{tab:new_labels}, with the 13 relations a \opTok{SUBST} link can name; relations under 1\,\% are pooled as other, and their shares are of \opTok{SUBST} links.}
\label{tab:new_labels_full}
\end{table*}

\section{Ablations}
\label{app:additional}
Table~\ref{tab:res_ablations} adds the components of our model one at a time and labels the same links in three ways.

\begin{table}[t]
\centering
\footnotesize
\setlength{\tabcolsep}{2pt}
\begin{tabular}{@{}lcccc@{}}
\toprule
\textbf{Row} & \textbf{macro F1} & \textbf{INFLECT} & \textbf{SUBST} & \textbf{$\Delta$} \\
\midrule
\multicolumn{5}{@{}l}{\textit{Components, added one at a time}} \\
Pointer, plain & \rv{abl.plain.l3} & \rv{abl.plain.infl} & \rv{abl.plain.subst} & n/a \\
+ evidence grid & \rv{abl.evidence.l3} & \rv{abl.evidence.infl} & \rv{abl.evidence.subst} & \rv{abl.evidence.delta} \\
+ stretch head & \rv{abl.stretch.l3} & \rv{abl.stretch.infl} & \rv{abl.stretch.subst} & \rv{abl.stretch.delta} \\
+ \opTok{COPY} gate & \rv{abl.gate.l3} & \rv{abl.gate.infl} & \rv{abl.gate.subst} & \rv{abl.gate.delta} \\
\midrule
\multicolumn{5}{@{}l}{\textit{The untyped pointer's links, labeled by}} \\
a trained head & \rv{abj.head.l3} & \rv{abj.head.infl} & \rv{abj.head.subst} & n/a \\
the rule typer & \rv{abj.rule.l3} & \rv{abj.rule.infl} & \rv{abj.rule.subst} & \rv{abj.rule.delta} \\
ours, no gate & \rv{abj.ours.l3} & \rv{abj.ours.infl} & \rv{abj.ours.subst} & \rv{abj.ours.delta} \\
\bottomrule
\end{tabular}
\caption{\textbf{Ablations of our model.} Mean over five folds, $\times100$: macro F1, \opTok{INFLECT} and \opTok{SUBST} F1, and $\Delta$, the paired gain in macro F1 over the row above with the folds won out of five. The \opTok{COPY} gate is the copy rule at decoding, and the last component row is our model. Restoring split links adds \rv{abl.split.delta}.}
\label{tab:res_ablations}
\end{table}

\section{Operation Profiles as Evidence of Reuse}
\label{app:downstream}
Table~\ref{tab:res_downstream} scores the classifier of Section~\ref{subsec:res_downstream} for every system, Table~\ref{tab:ana_features} defines its features, and Table~\ref{tab:ana_coef} gives their coefficients for each contrast.

The better the scripts in Table~\ref{tab:res_ac}, the better the evidence: the span aligner comes close to ours (per-query macro F1 \rv{ds.span_aligner.f1} against \rv{ds.ours.f1}), and the embedding aligners and the statistical aligner fall to the level of the similarity score.
The mean link confidence, the number of links, and the longest run carry the decision between a reference and a no-match pair (Table~\ref{tab:ana_coef}); no single operation decides it, since the label shares sum to one wherever a link exists and act together as an offset against the link count.
Split by kind, a verbatim reference is told from a no-match pair by how many words it takes and in how long a run, an allusion almost only by how confident its few links are, which is why allusions are the kind lost to no match.
Among the true references, the reuse words of a verbatim reference fall under the verbatim association where those of an allusion fall under no match, and the links of an allusion are more often inflections; the classifier weighs substitutions and reorderings toward allusion, but on their own they order the two kinds barely better than chance (AUC \rv{ana.auc.lex-rel-share} to \rv{ana.auc.subst-share}, Table~\ref{tab:ana_features}).

\begin{table}[t]
\centering
\small
\setlength{\tabcolsep}{3pt}
\begin{tabularx}{\columnwidth}{@{}Xccc@{}}
\toprule
 & \multicolumn{2}{c}{\textbf{Reference vs.\ no match}} & \textbf{\textit{cit.}/\textit{cf.}} \\
\cmidrule(lr){2-3}\cmidrule(lr){4-4}
\textbf{Script from} & \textbf{F1} & \textbf{R@10} & \textbf{F1} \\
\midrule
Annotation & n/a & n/a & \rv{ds.gold.type} \\
Ours & \rv{ds.ours.f1} & \rv{ds.ours.r10} & \rv{ds.ours.type} \\
Span aligner & \rv{ds.span_aligner.f1} & \rv{ds.span_aligner.r10} & \rv{ds.span_aligner.type} \\
Embedding similarity & \rv{ds.sim_aligner.f1} & \rv{ds.sim_aligner.r10} & \rv{ds.sim_aligner.type} \\
\quad fine-tuned & \rv{ds.sim_aligner_ft.f1} & \rv{ds.sim_aligner_ft.r10} & \rv{ds.sim_aligner_ft.type} \\
Statistical aligner & \rv{ds.em_aligner.f1} & \rv{ds.em_aligner.r10} & \rv{ds.em_aligner.type} \\
\midrule
Similarity score & \rv{ds.similarity.f1} & \rv{ds.similarity.r10} & n/a \\
Passage lengths & \rv{ds.lengths.f1} & \rv{ds.lengths.r10} & \rv{ds.lengths.type} \\
\bottomrule
\end{tabularx}
\caption{\textbf{Scripts as evidence of reuse.} A logistic regression over the script features separates references from no-match pairs (per-query macro F1, Recall@10; both reference types pooled) and verbatim references from allusions among the true references (macro F1). Mean over five folds, $\times100$. The last two rows use no script.}
\label{tab:res_downstream}
\end{table}

\begin{table*}[t]
\centering
\small
\begin{tabular*}{\textwidth}{@{\extracolsep{\fill}}llll@{}}
\toprule
\textbf{Feature} & \textbf{Meaning} & \textbf{Group} & \textbf{AUC} \\
\midrule
\AnalysisFeatureRows
\bottomrule
\end{tabular*}
\caption{\textbf{Script features of the downstream classifier.} Each of the \rv{ana.features} features is read off the decoded script of a candidate pair. AUC measures how well a feature alone ranks verbatim references above allusions among the true references (0.5 is chance), on our model's scripts over all five folds.}
\label{tab:ana_features}
\end{table*}

\begin{table*}[t]
\centering
\scriptsize
\setlength{\tabcolsep}{2pt}
\AnalysisCoefBlocks
\caption{\textbf{Feature coefficients per contrast.} Standardized coefficients of a logistic regression over our model's scripts, one contrast per row, averaged over five fits by test fold. Blue favors the first kind of pair of the row, orange the second, and the shade follows the size; $\dagger$ marks a feature whose sign differs between the fits. The first row is the reference classifier of Table~\ref{tab:res_downstream}, the two middle rows sample \rv{ana.negperpos} no-match pairs per positive as that classifier does, and the last uses the true references only. Table~\ref{tab:ana_features} defines the features.}
\label{tab:ana_coef}
\end{table*}

\section{Error Analysis}
\label{app:errors}
Table~\ref{tab:res_errors} sorts our model's errors on the two hard labels, on fold 4 and before the copy rule at decoding, by where the information is lost; Table~\ref{tab:res_bounds} corrects each kind of error in the predictions and shows what it costs.

\begin{table*}[t]
\begin{minipage}[t]{0.48\textwidth}
\centering
\footnotesize
\setlength{\tabcolsep}{3pt}
\begin{tabular}{@{}lcc@{}}
\toprule
\textbf{Error} & \textbf{INFLECT} & \textbf{SUBST} \\
\midrule
\multicolumn{3}{@{}l}{\textit{Missed}} \\
no link drawn & \rv{err.miss.nolink.infl} & \rv{err.miss.nolink.subst} \\
right link, wrong label & \rv{err.miss.label.infl} & \rv{err.miss.label.subst} \\
wrong source word & \rv{err.miss.source.infl} & \rv{err.miss.source.subst} \\
\midrule
\multicolumn{3}{@{}l}{\textit{False alarm}} \\
invented link & \rv{err.fa.invented.infl} & \rv{err.fa.invented.subst} \\
a \opTok{COPY} labeled as this & \rv{err.fa.copy.infl} & \rv{err.fa.copy.subst} \\
the other hard label & \rv{err.fa.other.infl} & \rv{err.fa.other.subst} \\
wrong source word & \rv{err.fa.source.infl} & \rv{err.fa.source.subst} \\
\bottomrule
\end{tabular}
\caption{\textbf{Errors on the hard labels.} Counts on fold 4 for our model before the copy rule at decoding.}
\label{tab:res_errors}
\end{minipage}\hfill
\begin{minipage}[t]{0.48\textwidth}
\centering
\footnotesize
\setlength{\tabcolsep}{2.5pt}
\begin{tabular}{@{}lccc@{}}
\toprule
\textbf{Predictions} & \textbf{SUBST} & \textbf{INFLECT} & \textbf{macro F1} \\
\midrule
As predicted & \rv{bnd.pred.subst} & \rv{bnd.pred.infl} & \rv{bnd.pred.l3} \\
Labels corrected & \rv{bnd.typing.subst} & \rv{bnd.typing.infl} & \rv{bnd.typing.l3} \\
Invented links removed & \rv{bnd.null.subst} & \rv{bnd.null.infl} & \rv{bnd.null.l3} \\
Both corrected & \rv{bnd.both.subst} & \rv{bnd.both.infl} & \rv{bnd.both.l3} \\
\midrule
Annotated links, rule typer & \rv{bnd.gold.subst} & \rv{bnd.gold.infl} & \rv{bnd.gold.l3} \\
\bottomrule
\end{tabular}
\caption{\textbf{Cost of each kind of error.} F1 ($\times100$) of our predictions on fold 4, with the copy rule at decoding, when the labels on the drawn links, the invented links, or both are corrected from the annotation; the last row names the annotated links with the shared rule.}
\label{tab:res_bounds}
\end{minipage}
\end{table*}

\clearpage

\end{document}